\pdfoutput=1

\documentclass[11pt]{article}

\usepackage[final]{latex/EMNLP2024}

\usepackage{times}
\usepackage{latexsym}

\usepackage[T1]{fontenc}
\usepackage{amsmath}
\usepackage{amssymb}
\usepackage{bbm}
\usepackage{subfig}
\usepackage{booktabs} 
\usepackage{multirow}
\usepackage{arydshln}
\usepackage{tabularray}
\usepackage[utf8]{inputenc}

\usepackage{microtype}

\usepackage{inconsolata}

\usepackage{graphicx}
\usepackage[font=small]{caption}

\title{On the Universal Truthfulness Hyperplane Inside LLMs}

\author{Junteng Liu$^{1}$, Shiqi Chen$^{2}$, Yu Cheng$^{3}$, Junxian He$^{1}$\\
  $^{1}$The Hong Kong University of Science and Technology, 
  $^{2}$City University of Hong Kong,\\
  $^{3}$The Chinese University of Hong Kong \\
  \texttt{jliugi@connect.ust.hk},
  \texttt{junxianh@cse.ust.hk} \\}

\begin{document}
\maketitle
\begin{abstract}
While large language models (LLMs) have demonstrated remarkable abilities across various fields, hallucination remains a significant challenge. 
Recent studies have explored hallucinations through the lens of internal representations, proposing mechanisms to decipher LLMs' adherence to facts. 
However, these approaches often fail to generalize to out-of-distribution data, leading to concerns about whether internal representation patterns reflect fundamental factual awareness, or only overfit spurious correlations on the specific datasets.
In this work, we investigate whether a universal truthfulness hyperplane that distinguishes the model's factually correct and incorrect outputs exists within the model.
To this end, we scale up the number of training datasets and conduct an extensive evaluation -- we train the truthfulness hyperplane on a diverse collection of over 40 datasets and examine its cross-task, cross-domain, and in-domain generalization.  
Our results indicate that increasing the diversity of the training datasets significantly enhances the performance in all scenarios, while the volume of data samples plays a less critical role. 
This finding supports the optimistic hypothesis that a universal truthfulness hyperplane may indeed exist within the model, offering promising directions for future research. Code is publicly available at \url{https://github.com/hkust-nlp/Universal\_Truthfulness\_Hyperplane}.
\end{abstract}

\section{Introduction}
\label{sec:intro}

Although large language models (LLMs) have gained significant success in a wide range of domains~\citep{openai2023gpt, llama, llama2}, hallucination problems remain the main challenges that hinder their wider applications~\citep{ji2023survey, zhang2023siren, huang2023survey}.
This issue is further aggravated by a limited understanding of the opaque inner mechanisms of LLMs' factual behaviors. 
Recent works start to investigate hallucinations from the perspective of inner representations, adopting the probing method~\citep{alain2016understanding} to identify hyperplanes in the space of hidden states to distinguish between correct responses and hallucinations~\citep{burns2022discovering,azaria2023internal,li2023inference,zou2023representation,marks2023geometry,ch2023androids}.
The underlying hypothesis is that the hidden states of language models already encode significant information on hallucination, and we are able to tell hallucinations from the hidden states.




While these studies have achieved impressive hallucination detection performance on the datasets which the probes are trained on~\citep{burns2022discovering,li2023inference,zou2023representation,marks2023geometry,ch2023androids}, they often struggle to 
generalize to out-of-distribution (OOD) data samples~\citep{burns2022discovering,marks2023geometry,ch2023androids}.
We further verify such OOD generalization failure in our experiments, confirming that the performance of the probe trained solely on TruthfulQA~~\citep{lin2021truthfulqa} -- a widely used dataset to train probes~\citep{li2023inference,chen2023truth,joshi2023personas} -- will drop 25 absolute points on average for several other datasets compared to in-domain detection.
This failure raises two principled questions: (1) Do the identified inner representation features in previous works really capture the model's inner hallucination, or only overfit spurious patterns of the specific dataset? (2) Does there exist a \emph{universal} truthfulness hyperplane that can classify factual correctness on diverse tasks?



\begin{figure*} [t]
    \centering
    \includegraphics[width=0.85\textwidth]
    {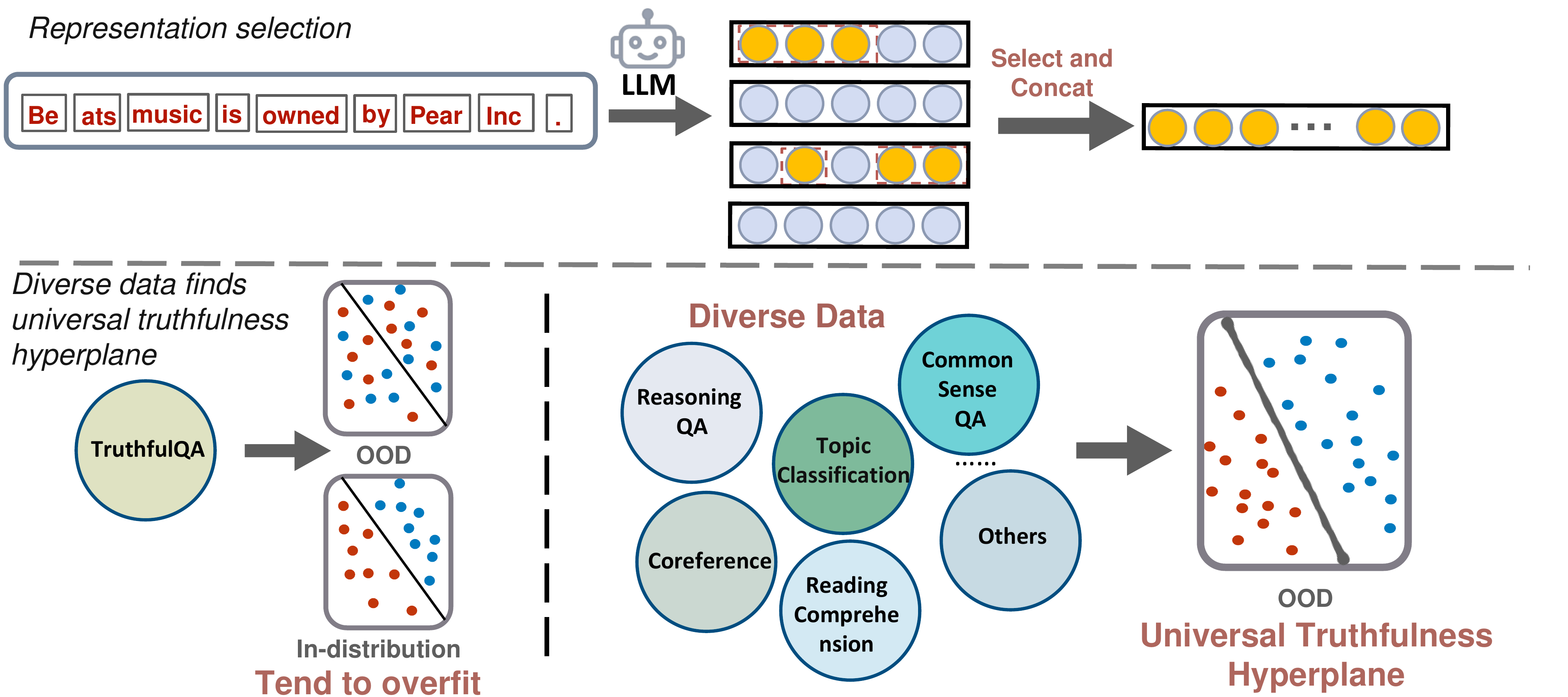}
    \vspace{-5pt}
    \caption{{\bf Top}: we extract representations from the last token of the input sequence, then specific locations of the hidden states inside the LLM are selected and concatenated as input to train the probe.
    {\bf Bottom}: Previous works mainly train the linear probe on one dataset which tends to overfit spurious features. Our work utilizes diverse datasets to examine whether a universal truthfulness hyperplane exists that can generalize to out-of-domain data. 
    }
    \vspace{-15pt}
    \label{fig-method}
\end{figure*}

\begin{figure*}[t]
    \centering
    \includegraphics[width=0.9\linewidth]{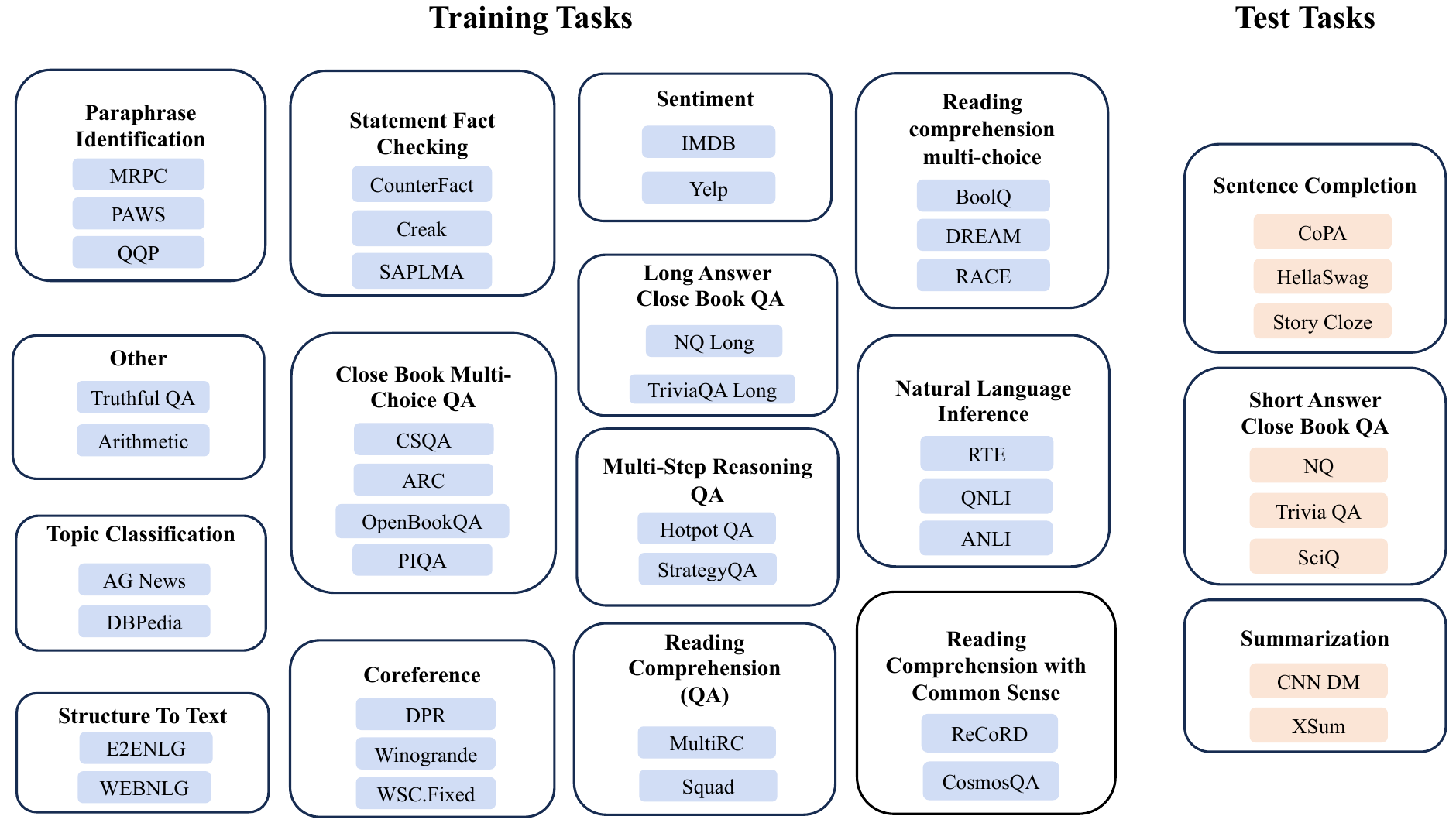}
    \vspace{-5pt}
    \caption{Our curated datasets and tasks. Left (Blue) part represents the training tasks, while the right (Orange) represents the test tasks.}
    
    \label{fig-dataset}
    \vspace{-10pt}
\end{figure*}

We aim to answer these questions in this work.
Inspired by the success of diversified instruction tuning \citep{sanh2021multitask,wei2021finetuned,chung2022scaling,wang-etal-2023-self-instruct}, our idea is to increase the diversity of the training data by scaling up the number of training datasets, so that we may find the universal truthfulness hyperplane that can generalize across tasks using the framework shown in Figure~\ref{fig-method}.
Specifically,
we construct a comprehensive and diverse collection of hallucination detection datasets to facilitate the analysis. 
The dataset comprises 17 distinct categories of tasks
covering over 40 datasets from knowledge-seeking QA tasks such as Triviaqa~\citep{2017arXivtriviaqa}, Natural Questions~\citep{kwiatkowski2019natural} to structure-to-text tasks such as E2ENLG~\citep{duvsek2020evaluating}, with each task containing both correct and incorrect samples, as illustrated in Figure \ref{fig-dataset}. 
These datasets enable us to thoroughly evaluate the performance and robustness of the truthfulness probes.

In our experiments, we train probes using diverse datasets and evaluate their generalization performance in three scenarios: cross-task, cross-domain, and in-domain.
We study the effectiveness of probing different locations of hidden states and find that the attention heads lead to the highest accuracy.
Our probe method beats the prompting-based approach as well as the probability baseline significantly and outperforms the previous probe that is trained only on one dataset by 14 absolute points, achieving $\sim$ 70\% cross-task accuracy.
This provides empirical evidence for the existence of a shared representation of truthfulness within the model.
Notably, despite our probe being trained on an extensive collection of datasets, it achieves high performance with an average of only 10 data samples per dataset. This demonstrates the method's data efficiency and its straightforward applicability in identifying a universal truthfulness hyperplane.

\section{Probing Hidden States for Truthfulness}
\label{sec:data}

\subsection{Overview}
\label{sec:overview}
Probing methods are defined as training classifiers with hidden states of the neural networks as input to identify specific properties of the input~\citep{alain2016understanding,belinkov-2022-probing}. Previous works primarily focus on the linguistic information in representations~\citep{jawahar-etal-2019-bert,tenney2018what}, while recent works explore truthfulness as the property and design probes to detect the truthfulness of large language models~\citep{li2023inference,chen2023truth,marks2023geometry,zou2023representation,ch2023androids}.
In addition to typical linear supervised probes like logistic regression (LR)~\citep{ch2023androids} and mass mean (MM)~\citep{marks2023geometry}, unsupervised linear probes such as CCS~\citep{burns2022discovering} and LAT~\citep{zou2023representation} are also studied for truthfulness. 
Previous works train the probe exclusively on one or a few specific datasets and subsequently evaluate their performance on the same or similar datasets~\citep{li2023inference,chen2023truth,azaria2023internal,marks2023geometry}, which may overfit the spurious features of the datasets and fail to capture the underlying truthfulness inside the model. In contrast, our objective in this work is to examine the existence of a \emph{universal} truthfulness hyperplane encoded in the trained probes that can generalize well across various datasets.

\subsection{Formulation}
\label{sec:form}
As many works argue that the linear representations for high-level semantic concepts in LLMs~\citep{tigges2024language,jiang2024origins} and the linear structure probes offer good interpretability,
we employ two linear probing methods: logistic regression (LR) and mass mean (MM) to extract truthfulness from the hidden states of LLMs in this paper. 
Formally, given a dataset $D=\{(x_i,y_i) | i=1,\cdots, N\}$, where $x_i$ is a data sample and $y_i\in \{0,1\}$ indicates whether $x_i$ is factually correct or not, we extract the representations by $h_i=\phi(x_i)$ and then categorize them into two parts: $H^{+}=\{h_i|y_i=1\}$ and $H^{-}=\{h_i|y_i=0\}$. As $x_i$ is a text sequence in our context, we compute $h_i$ as the representation of the last token in $x_i$ from a transformer model~\citep{vaswani2017attention} across this paper, and in \textsection\ref{sec:probe} we will discuss the specific hidden states locations (e.g., from which layer to extract $h_i$) from transformers to extract $h_i$. The LR and MM probes learn different truthfulness vectors:
 \begin{equation}
\begin{aligned}
\theta_{\text{lr}} = \arg\min_{\theta}  \sum_{i} \left[ y_i\log\left(\sigma(\theta^{T} h_i)\right) + \right.  \\ 
\left. (1-y_i)\log\left(1-\sigma(\theta^{T} h_i)\right) \right] ,
\end{aligned}
\end{equation}

\begin{equation}
\theta_{\text{mm}}= \overline{H^{+}} - \overline{H^{-}},
\end{equation}
where $\overline{H^{+}}$ and $\overline{H^{-}}$ correspond to the average representations of the sets $H^{+}$ and $H^{-}$, respectively. 
$\theta_{\text{lr}}$ is from logistic regression and $\theta_{\text{mm}}$ just aligns with the direction from $\overline{H^{-}}$ to $\overline{H^{+}}$.
After obtaining $\theta$, classification is performed as $y_i=\mathbbm{1}(\theta^Th\geq0)$ where $\mathbbm{1}$ is the indicator function. 
This way, $\theta^{T}h=0$ essentially defines a linear hyperplane that is orthogonal to the direction of the truthful vector $\theta$ in the space of $h$ to classify truthfulness, and we refer to it as the \emph{truthfulness hyperplane}. The truthfulness hyperplane may be specific to datasets, or universal across different distributions that represent the self-awareness of the truthfulness of the model, which is the question we aim to study in this work.


\subsection{Data Curation}
Previous probing papers all focus on training the probes exclusively on one or one type of dataset so that they may fail to obtain the universal truthfulness hyperplane and overfit to the specific data. For example,~\citet{li2023inference,chen2023truth} primarily train and evaluate on TruthfulQA~\citep{lin2021truthfulqa}, while~\citet{azaria2023internal,marks2023geometry} mainly concentrate on datasets containing single-sentence true or false statements. Meanwhile,~\citet{ch2023androids} only consider the truthfulness probe on in-context generation tasks. Some works have observed the failure of generalization in OOD data samples~\citep{burns2022discovering,marks2023geometry,ch2023androids}. Our experiments of OOD generalization failure  of probes solely trained on TruthfulQA in ~\textsection\ref{ood-failure} further validate that the learned hyperplane in the probe is overfitting on the trained distribution and not universal.

Therefore, to find the potentially universal truthfulness hyperplane,
we create and collect a variety of datasets used for hallucination detection. Following the task taxonomy from T0 and Flan \citep{sanh2021multitask, wei2021finetuned}, we create a collection of 49 datasets in 17 tasks,\footnote{The term `task' is used to refer to a group of similar datasets.} shown in Figure~\ref{fig-dataset}. 
We aim to conduct hallucination detection that requires both correct and incorrect data. To collect incorrect data points, for datasets that pair with false answers, such as multiple-choice questions, we select the wrong answers randomly as the responses. 
For text generation tasks that typically only consist of a single correct answer, we employ two different strategies to produce incorrect data examples: For the grounding-based text generation dataset E2ENLG~\citep{duvsek2020evaluating}, we randomly replace attributes to produce false attributes. Meanwhile, we utilize the GPT-3.5-turbo for WEBNLG \citep{gardent2017creating} and the GPT-4-turbo for other datasets (e.g.~TriviaQA \citep{2017arXivtriviaqa}), to generate convincing but false answers.

As shown in Figure~\ref{fig-dataset}, we split the tasks into training tasks and test tasks to evaluate cross-task generalization.
For each dataset, we use a prompt template to format the input and divide the dataset into training, validation, and test splits. 
It is important to note that the training split for every dataset consists of up to 800 data samples and each validation split has 100 data samples, while the remaining samples are used as the test splits. 
We find that 800 training samples for each dataset are enough to train the probe and we do not observe significant gains as we further increase the training samples, as we will show in \textsection\ref{analysis}.
More details on data curation are discussed in Appendix~\ref{sec:appendix}.

\subsection{The Probe Design}
\label{sec:probe}

\paragraph{Input Representations:}
In \textsection\ref{sec:form} we have described to use the representation of the last token of the input sequence as the feature $h$. 
The last-token representation is commonly used as sentence representations as it aggregates the overall sentence information~\citep{burns2022discovering,li2023inference}.
However, the specific locations inside the transformer model to extract the representations are still up to decide -- for example, which layer of hidden states to use? Shall we use attention activation or layer residual activation?  
Various previous studies have explored probing on different types of representations.~\citet{li2023inference,campbell2023localizing} conduct truthfulness probing on the attention head outputs, another line of works considers using the layer residual activations~\citep{burns2022discovering,azaria2023internal,marks2023geometry}.
Among these works,~\citet{burns2022discovering} select the last layer residual activation as input to train probes, while ~\citet{azaria2023internal,marks2023geometry} utilize specific intermediate layers to train probes.
Based on our preliminary experiments, we determine that attention head outputs serve as an effective representation, denoted as $h$, for training our probe. We will report the ablation results in \textsection\ref{head-vs-layer} to compare the attention head outputs to the layer residual stream activations.
Besides, one layer, or especially one attention head may not be expressive enough, and the truthfulness inside the model may be captured by different locations of representations together. 
Therefore, we consider combining the attention heads across different layers. 
Relevantly, \citet{ch2023androids} train probes in each layer respectively and ensemble all of them to make the final prediction. However, we argue that using all hidden states inside the model results in significant redundancy during training and inference time, and it is likely that only a small fraction of the hidden states capture the truthfulness information.  
Therefore, we adopt a hidden states location selection strategy to select and combine certain representations of the last token in the input sequence to train the probe, as we detail next. An overview of the input feature extraction is illustrated in Figure~\ref{fig-method}.  


\paragraph{Selecting Hidden States Locations:}
We hypothesize that only a small fraction of the representations in the transformer model is related to truthfulness, 
and within these hidden states, different locations may contain varying information about the truthfulness of diverse datasets or different aspects of the same dataset. 
Therefore, we perform a preliminary probe training procedure to select the specific locations of representations of the last token.
Concretely, we train a preliminary probe for each attention head across all layers of the last token respectively on the aggregated training splits of the training tasks, which leads to 1024 (32 layers x 32 heads) different probes based on LLaMA2-7b-chat~\citep{llama2} representations. 
Then we measure the truthfulness classification accuracies of these probe models on the validation split of each dataset in the training tasks respectively. 
Subsequently, for each validation split, we select the top $num$ locations with the highest accuracy. 
Such a procedure will select out at most $41*num$ locations in total after removing duplicates where 41 is the number of validation splits.
Finally, we concatenate the representations of all these selected locations as the input to train the final probe model. $num$ is a tunable hyperparameter and we find that larger $num$ does not always produce better results -- in fact, in our experiments a $num$ equal to 1 or 2 typically yields the best performance. 
We include the ablation results on $num$ in Appendix~\ref{sec:appendix-hyperpara-num}.

\begin{figure*} [t]
	\centering
	\subfloat[\label{fig-choices-a}\vspace{-10pt}Trivia QA]{
	\hspace{-0.4cm}	
  \includegraphics[scale=0.3]{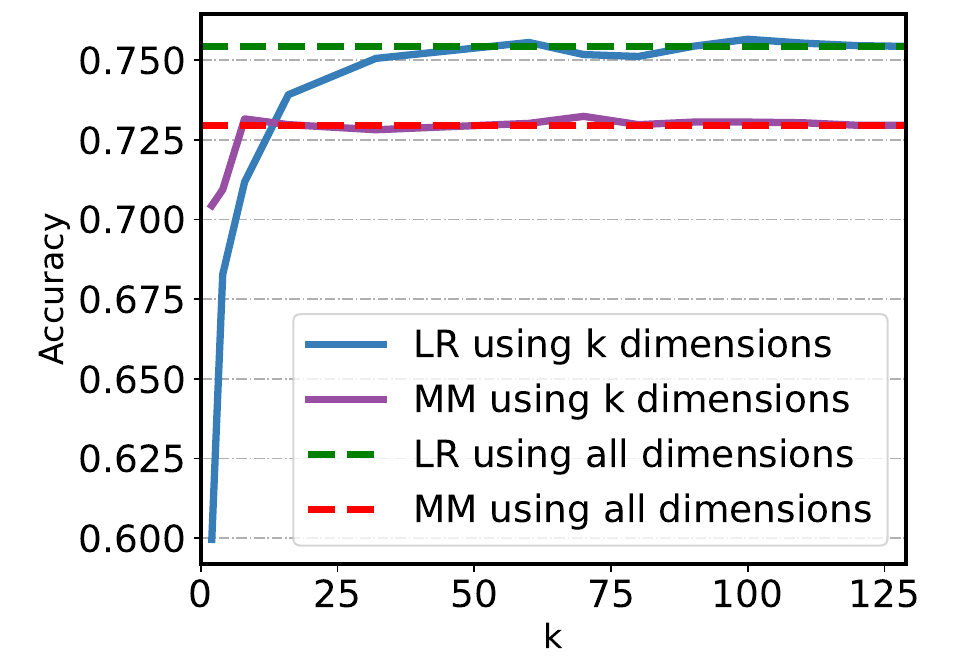}}
	\subfloat[\label{fig-choices-b}XSum]{
    \hspace{-0.38cm}
		\includegraphics[scale=0.3]
  {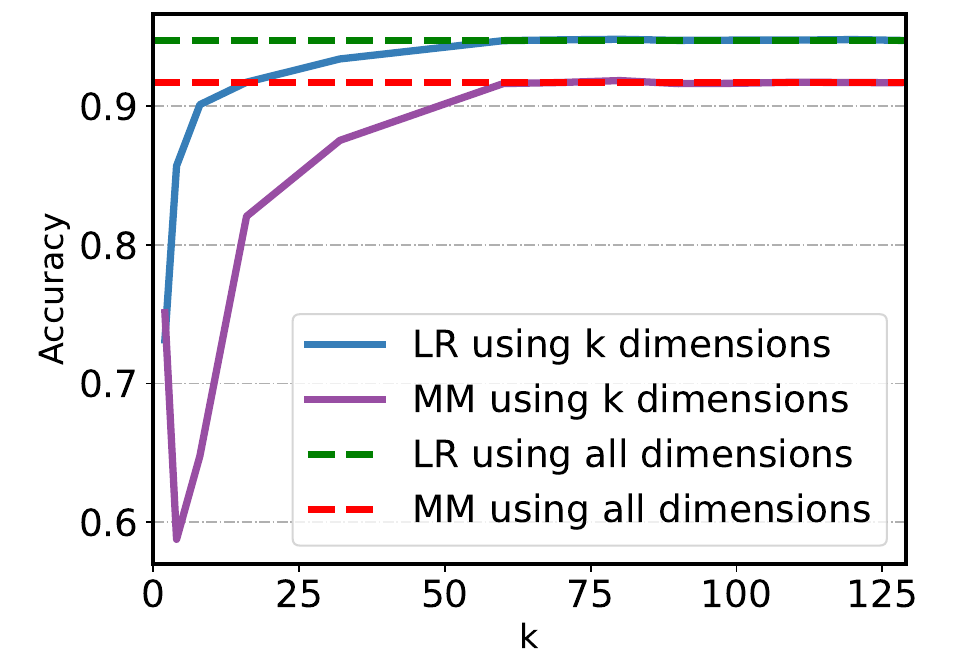} }
	\subfloat[\label{fig-choices-c}HellaSwag]{
        \hspace{-0.42cm}
		\includegraphics[scale=0.3]{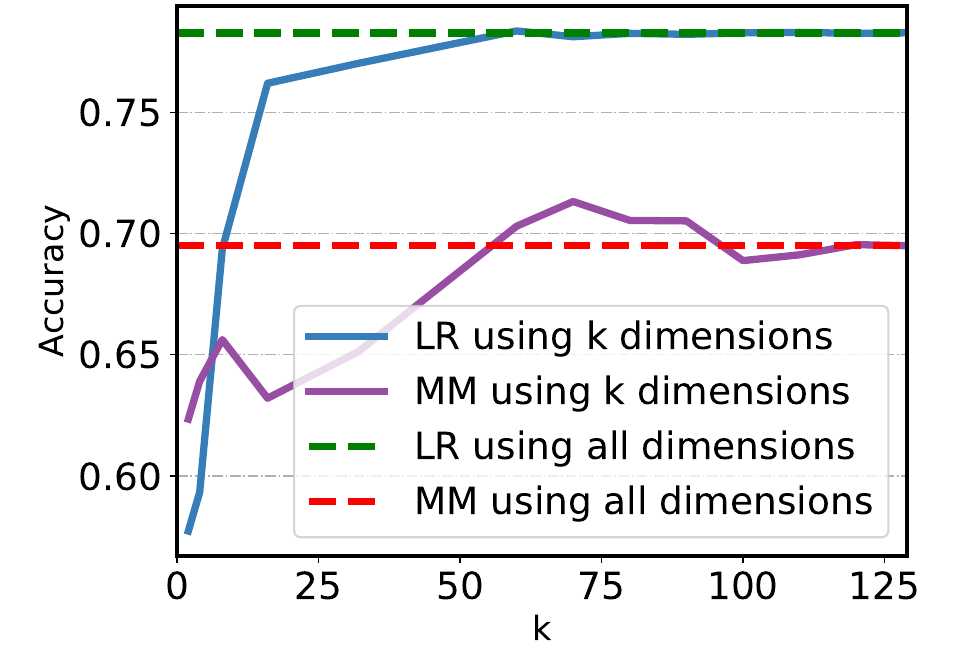} }
 \vspace{-10pt}
	\caption{Examples of sparsity test on different datasets using the logistic regression (LR) and the mass mean (MM) probe.
 }
  \vspace{-10pt}
	\label{fig-sparsity-show} 
\end{figure*}

\label{compression}

\paragraph{Sparsity of Truthfulness Features:}
Even though we select only a small fraction of hidden representations of the model, the overall input features are still high-dimensional.
Inspired by \citet{sparse}, which examines the sparsity of learned representations by k-sparse probes trained on over 100 distinct features, we consider enforcing sparsity constraints in our probe design. 
Specifically, we evaluate the sparsity of truthfulness by employing the linear ranking method that utilizes the weights of the trained classifiers to rank the neurons and selects those with high ranks~\citep{dalvi2019one} -- we identify the indices of the largest $k$ values in $|\theta|$, then we index the corresponding $k$ features from the original $h$ to form the new input feature.
Our preliminary sparsity test, conducted on a single dataset and one attention head output, demonstrates that reducing the number of neurons by nearly half does not decrease task performance, as shown in Figure~\ref{fig-sparsity-show}, where the experiment details can be found in Appendix~\ref{sec:appendix-sparsity}. Consequently, we introduce this tunable hyperparameter $k$ used to compress each representation into $k$ dimensions. The hyperparameter $k$ can be set as 64 or 128, with 128 representing the full dimensionality of the attention head output for our used 3 models: LLaMA2-7b-chat~\citep{llama2}, LLaMA2-13b-chat~\citep{llama2} and Mistral-7b~\citep{jiang2023mistral}.

\section{Experiment}
\label{sec:experiment}
\subsection{General Setup}

We experiment under three evaluation settings: cross-task, cross-domain, and in-domain. In each setting, we evaluate on the same test tasks (3 tasks: sentence completion, short answer close book QA and summarization tasks, 8 datasets) shown in Figure~\ref{fig-dataset}. For a given value of the  hyperparameter $num$, we always adopt the validation splits of the training tasks as validation data for selecting $num$ positions. Concretely, for (1)
\textbf{Cross-Task}, the training data are the training splits of the training tasks; (2) \textbf{Cross-Domain},
the training data include the training splits of all the training tasks plus all the datasets within the current test task, except for the test dataset itself; 
and (3) \textbf{In-Domain}, we utilize the training splits of all the datasets -- including the training split of the test dataset itself -- to train the probe.
Generally, we emphasize the cross-task results the most, which we think reflects whether the learned hyperplane can generalize in the wild and is universal. We mainly conduct our experiments with the LLaMA2-7b-chat model \citep{llama2}, while in \textsection\ref{exp-other-model} we experiment with the Mistral-7b-v0.1 base model~\citep{jiang2023mistral} and the LLaMA2-13b-chat model~\citep{llama2} as well.
More details on the setup 
can be found in Appendix~\ref{sec:appendix-experiment}. 

\paragraph{Hyperparameters:}
There are two hyperparameters to tune in our probe model, $num$, which decides the number of representations to the input, and $k$ which denotes the compressed dimensions for every representation as indicated in \textsection\ref{sec:probe}. Hyperparameter tuning of $num$ and $k$ is performed exclusively on the test splits of the training tasks in Figure~\ref{fig-dataset}, ensuring that we never use the validation or test splits of our test tasks to select the hyperparameters. Please see Appendix~\ref{sec:appendix-hyperpara-tune} for details on hyperparameter tuning.


\paragraph{Baselines:}
We mainly compare our probe method with two baselines.
\textbf{(1) Self-Eval~\citep{kadavath2022language}:} In this approach, we directly prompt the model to assess the correctness of each data sample by the prompt such as ``Is the answer correct or wrong?''. Then we constrain the model to decode only from ``correct'' or ``wrong'' tokens. 
\textbf{(2) Probability:} 
This method calculates the probability of answers in data samples. In cases where the datasets contain long answers, such as TruthfulQA~\citep{lin2021truthfulqa} and E2ENLG~\citep{duvsek2020evaluating}, we normalize log probability by length to compute the per-token log probability. 
We classify the example to be factually correct when the probability is larger than a threshold $\tau$, which is a hyperparameter that is tuned on different training splits. Specifically, these splits are from datasets of different tasks for cross-task settings, a randomly different dataset within the same task for cross-domain settings, and the same dataset for in-domain settings.
For both Self-Eval and Probability baselines, we select data samples from these different training splits in the three settings as few-shot demonstrations.
In addition to the baselines, we also report results from the Finetuning method, where we fine-tune the entire model on the same training data as our probe to judge the truthfulness of the data sample. 
We note that the Finetuning method approximately serves as an upper bound of our probe method. This is because our work aims to identify the potentially universal truthfulness hyperplane where we do not change the model parameters or hidden states, while finetuning the models is given much more flexibility by updating the models. 

\subsection{Dedicated Probes Fail to Generalize}
\begin{table}[t]
    \centering
    \small
    \begin{tabular}{l c c}
        \toprule
        \textbf{Method} & \textbf{In-distribution} & \textbf{Average OOD} \\
        \midrule
        Random & 50.00 & 50.00 \\
        FT & 79.50 & 56.51 \\

        Self-Eval & 62.96 & \textbf{63.31} \\
        Probability & 55.96 & -- \\
        Probe-LR & \textbf{82.28} & 54.44 \\
        Probe-MM & 77.08 & 50.71 \\
        
        \bottomrule
    \end{tabular}
    \vspace{-5pt}
    \caption{The in-distribution and OOD accuracy of different probes trained on TruthfulQA, Self-Eval, Probability, and FT (finetuning) method (\%).  
    }
    \vspace{-15pt}
    \label{tab:tqa-id}
\end{table}

\label{ood-failure}

Before discussing the main results of our probe model, we first reproduce the settings in previous works where we train our probe model on the TruthfulQA dataset~\citep{lin2021truthfulqa,chen2023truth}.
TruthfulQA is a popular dataset measuring the truthfulness of models, and many works conduct truthfulness probing trained on TruthfulQA and are dedicated to improving the TruthfulQA performance~\citep{li2023inference,chen2023truth}. 
It is unknown whether the linear probes from previous works identify the real truthfulness hyperplane, or only overfitting to the truthfulness features of the TruthfulQA dataset.
Specifically, we train the probe on TruthfulQA and utilize the TruthfulQA validation split to tune the hyperparameters. We evaluate the probe on the TruthfulQA test split as in-distribution test, as well as 8 other datasets as out-of-distribution (OOD) test, which are from the test tasks in Figure~\ref{fig-dataset}. We report the average results, while the details of baselines and OOD results for every dataset can be seen in Appendix~\ref{detail-fail}.

\begin{table*}[t] 
\centering
\Large
\resizebox{1.5\columnwidth}{!}{
\begin{tabular}{ll ccc cc ccc c}
\toprule
\multicolumn{2}{l}{\multirow{2}{*}{\textbf{Method}}} & \multicolumn{3}{c}{\textbf{Short Answer Close Book QA}} &
\multicolumn{2}{c}{\textbf{Summarization}} &
\multicolumn{3}{c}{\textbf{Sentence Completion}} & \multirow{2}{*}{\textbf{Average}} \\
& & NQ & Trivia QA & SciQ & XSum & CNN DM & SC & HS & CoPA &  \\
\midrule
\multirow{5}{*}{\textbf{Cross-task}}  & FT & 69.92 & 73.34 & 80.00 & 78.66 & 85.68 & 72.07 & 73.68 & 88.00 & 77.67\\ \cdashline{2-11}
 & Self-Eval & 56.80 & 69.90 & 81.70 & 67.00 & 65.98 & 65.71 & 56.48 & 54.50 & 64.76\\
 & Probability & 57.56 & 68.96 & 68.05 & 52.12 & 61.94 & 56.95 &	49.30 & 72.50 & 60.92 \\
& Probe-LR& 63.90 & 71.36 & 76.90 & 63.98 & 80.66 & 70.71 & 64.40 & 62.00 & 69.24\\
& Probe-MM& 58.52 & 71.88 & 82.60 & 75.82 & 71.38 & 73.06 & 59.50 & 71.00 & \textbf{70.47}\\
 \midrule
 \multirow{5}{*}{\textbf{Cross-domain}}  & FT &  70.54 & 73.54 &	80.70 & 58.20	 & 95.82 & 71.43 & 73.18 & 85.50 & 76.11\\ \cdashline{2-11}
  & Self-Eval & 56.78 & 68.92 & 81.55 & 67.00 & 65.98 & 67.40 & 61.52 & 59.50 & 66.08 \\
 & Probability & 57.18 &	67.72 & 65.70 & 53.50	& 58.04	& 68.15 & 49.24 & 81.00 & 62.57 \\
 
 & Probe-LR & 64.66 & 71.48 & 79.45 & 65.64 & 85.34 & 68.79 & 67.06 &  68.50 & \textbf{71.36} \\
  & Probe-MM & 58.64 & 71.82 & 82.80 & 67.66 & 73.22 & 72.80 & 63.60 &  65.50 & 69.50 \\

 \midrule
\multirow{5}{*}{\textbf{In-domain}} & FT & 70.16 & 76.80 & 83.85 & 96.20 & 99.38 & 74.27 & 87.38 & 93.50 & 85.19\\ \cdashline{2-11}
& Self-Eval & 57.60 & 70.96 & 84.30 & 67.00 & 65.98	& 66.92 & 58.04	 & 78.50	& 68.66 \\
& Probability & 56.66 & 70.54 & 85.20  & 54.46 & 62.52 & 69.70 & 52.68 & 88.50	& 67.53 \\
& Probe-LR & 67.34 & 74.50 & 82.80 & 90.20 & 95.88 &	72.98 &	73.80 &  75.00& \textbf{79.06} \\
& Probe-MM & 58.56 & 71.96 & 83.55 & 78.08 & 76.88 &	72.47 &	61.12 &  70.50 & 71.64 \\
\bottomrule
\end{tabular}
}
 \vspace{-5pt}
\caption{Results of training on diverse datasets, where FT indicates the Finetuning method, SC indicates the Story Cloze dataset, and HS indicates the HellaSwag dataset.}
\label{tab:main-result}
 \vspace{-10pt}
\end{table*}

\paragraph{Results:}

The in-distribution and out-of-distribution (OOD) performance are reported in Table~\ref{tab:tqa-id}. For OOD evaluation, we present the average accuracy across the test tasks.  
Our findings indicate that in the in-distribution TruthfulQA test, the probe method surpasses both the Self-Eval and Probability baselines by nearly 20 percentage points. In stark contrast, the probe method's performance deteriorates significantly when tested on OOD data, lagging behind the Self-Eval baseline by approximately 10 percentage points. The probe's accuracy, close to the chance level at 50, implies that the learned hyperplane of the probe fails to contain any truthfulness information pertinent to certain OOD datasets.
This OOD generalization failure observation is consistent with prior research \citep{ch2023androids, marks2023geometry}, which suggests that representations of truthfulness are highly task-specific and distribution-dependent. The failure underscores that the hyperplane derived from training solely on the TruthfulQA dataset is not the universal truthfulness hyperplane.

\subsection{Main Results -- On the Universal Truthfulness Hyperplane}
\label{main-exp}

To investigate the existence of the universal truthfulness hyperplane, we report the results of both the logistic regression probe (Probe-LR) and the mass mean probe (Probe-MM) in the cross-task, cross-domain, and in-domain settings respectively. Descriptions of the two probes can be found in \S\ref{sec:overview}.
In Table~\ref{tab:main-result}, we observe that both Probe-LR and Probe-MM consistently outperform the Self-Eval and Probability baselines across all three settings, with average improvements of 5.10, 4.35, 6.69 absolute percentage points respectively over the stronger baseline. The Probe-MM method outperforms the two baselines on 7 out of 8 test datasets in the cross-task setting. 
~\textbf{Notably, both probe methods achieved approximately 70\% accuracy in the challenging cross-task setting}. 
Compared to previous OOD generalization failure, our results convey positive signals on the existence of a universal truthfulness hyperplane inside LLMs. 
Comparing Probe-LR to Probe-MM, Probe-LR outperforms Probe-MM in both cross-domain and in-domain settings, while Probe-MM  exhibits slightly better generalization performance in the cross-task scenario, which is expected since the Probe-MM does not specifically ``train'' the classifier through optimization, thus less likely to overfit to spurious patterns of the training data, similar findings have been presented before in~\citet{marks2023geometry}. 
Notably, Finetuning (FT) achieves the highest accuracy, reaching over 75\% accuracy across all three settings. 
These results demonstrate the practicality of FT on this task, and imply that a well-tuned model may be able to classify truthfulness reasonably well.
However, we note that Finetuning neither produces any interpretation on the hidden states of the model, nor answers our central question on whether a universal truthfulness hyperplane exists of not. 
We emphasize our focus of this work on exploring whether LLMs' hidden states express the inner notion of truthfulness in a simple way, i.e., with a linear hyperplane.

\subsection{Experiments on Other Models}
\label{exp-other-model}

We also explore our method in the Mistral-7b-v0.1 base model~\citep{jiang2023mistral} and the LLaMA2-13b-chat model~\citep{llama2}, conducting cross-task experiments. The results are shown in Table~\ref{tab:other-models}.
Consistent with the findings from the LLaMA2-7b-chat experiments, Probe-MM demonstrates superior generalization compared to Probe-LR, particularly for the Mistral-7b model. Specifically, Probe-MM achieves better performance than both the Self-Eval and Probability baselines for both models, exhibiting a substantial improvement of 12.81 absolute points for Mistral-7b and 1.23 points for LLaMA2-13b-chat. Moreover, Probe-MM outperforms the baselines on 7 out of 8 datasets for Mistral-7b and 5 out of 8 datasets for LLaMA2-13b-chat.
Notably, both Mistral-7b and LLaMA2-13b-chat achieve higher cross-task accuracies than LLaMA-7b-chat in Table~\ref{tab:main-result}, with Mistral-7b reaching 77.11 and LLaMA2-13b-chat reaching 73.88, revealing a positive trend that the universal truthfulness hyperplane within the hidden states of more advanced LLMs tends to become more pronounced. The details for hyperparameter tuning can be seen in Appendix~\ref{sec:appendix-hyperpara-tune}.

\begin{table*}[t] 
\centering
\Large
\resizebox{1.5\columnwidth}{!}{
\begin{tabular}{ll ccc cc ccc c}
\toprule
\multicolumn{1}{l}{\multirow{2}{*}{\textbf{Model}}}  & \multirow{2}{*}{\textbf{Method}} & \multicolumn{3}{c}{\textbf{Short Answer Close Book QA}} &
\multicolumn{2}{c}{\textbf{Summarization}} &
\multicolumn{3}{c}{\textbf{Sentence Completion}} & \multirow{2}{*}{\textbf{Average}} \\
& & NQ & Trivia QA & SciQ & XSum & CNN DM & SC & HS & CoPA &  \\
\midrule
\multirow{4}{*}{\textbf{Mistral-7b}}

 & Self-Eval & 60.44 & 66.08 & 79.35 & 61.34 & 52.96 & 51.84 & 50.76 & 50.00 & 59.10\\
 & Probability & 61.00 & 74.34 & 60.45 & 56.36 & 57.04 & 66.81 &	50.40 & 88.00 & 64.30 \\
  & Probe-LR& 67.10 & 78.08 & 78.60 & 75.90 & 76.30 & 68.95 & 59.76 & 72.50 & 72.15\\
& Probe-MM& 63.84 & 77.56 & 87.35 & 84.60 & 81.74 & 71.75 & 69.00 & 81.00 & \textbf{77.11}\\
\midrule
\multirow{4}{*}{\textbf{LLaMA2-13b-chat}} 
 & Self-Eval & 59.14 & 71.52 & 83.40 & 76.94 & 80.60 & 68.92 & 61.48 & 83.50 & 72.65\\
 & Probability & 61.90 & 72.34 & 74.70 & 54.60 & 61.14 & 70.34 & 49.36 & 84.50 & 66.11 \\
 & Probe-LR& 66.88 & 76.40 & 79.50 & 72.22 & 84.50 & 72.31 & 56.24 & 72.50 & 72.57\\
& Probe-MM& 59.74 & 74.62 & 85.80 & 71.66 & 81.54 & 71.14 & 67.04 & 79.50 & \textbf{73.88}\\
\bottomrule
\end{tabular}
}
 \vspace{-5pt}
\caption{The result of cross-task experiments on Mistral-7b and LLaMA2-13b-chat models, where SC indicates the Story Cloze dataset, and HS indicates the HellaSwag dataset.}
\vspace{-15pt}
\label{tab:other-models}
\end{table*}

\begin{figure*} [t]
	\centering
     \subfloat[\label{fig-atten-layer}The average cross-task accuracy of different probes trained using attention head outputs and layer
residual activations on varying datasets.]{
    \includegraphics[scale=0.26]{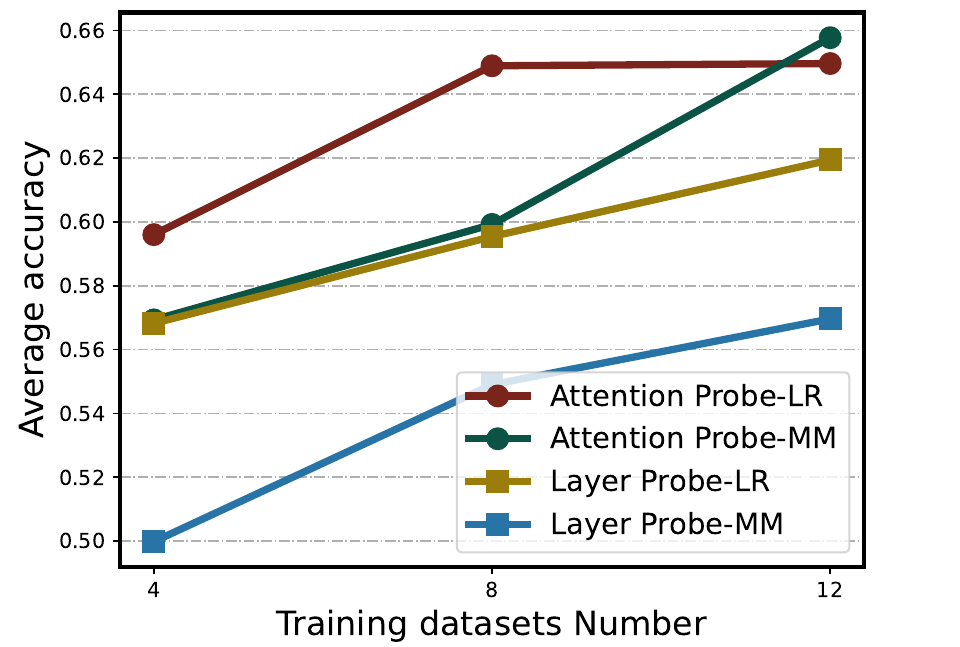}}
    \hspace{0.3cm}
	\subfloat[\label{fig-scaling}The average cross-task accuracy of different probes and FT trained on scaling number of training tasks.]{
		\includegraphics[scale=0.26]{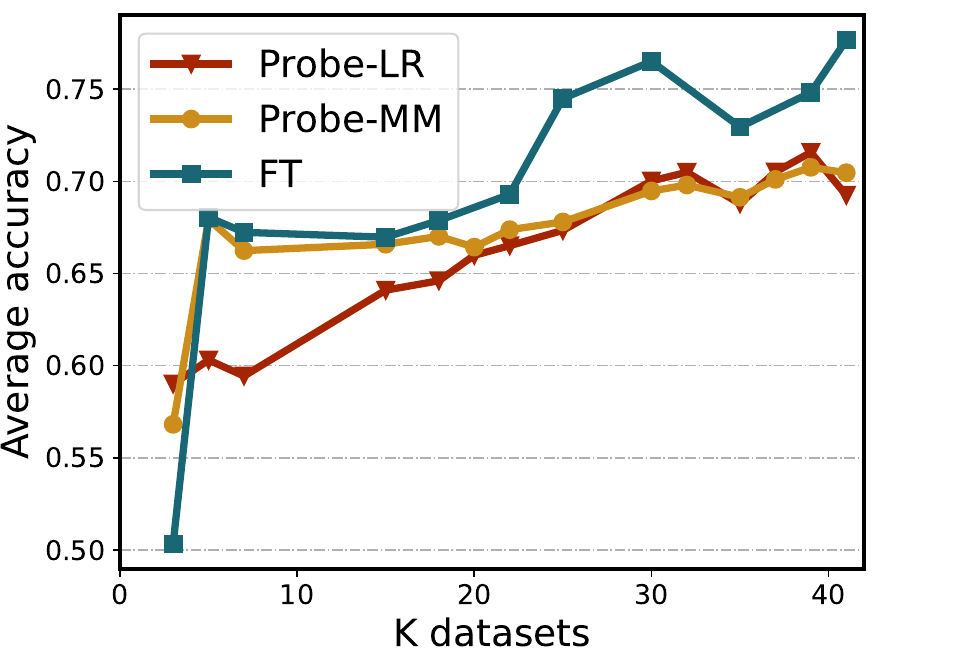}}
	\hspace{0.3cm}
	\subfloat[\label{fig-scal-num}The average cross-task accuracy of different probes trained on varying training split size per dataset.]{
		\includegraphics[scale=0.26]{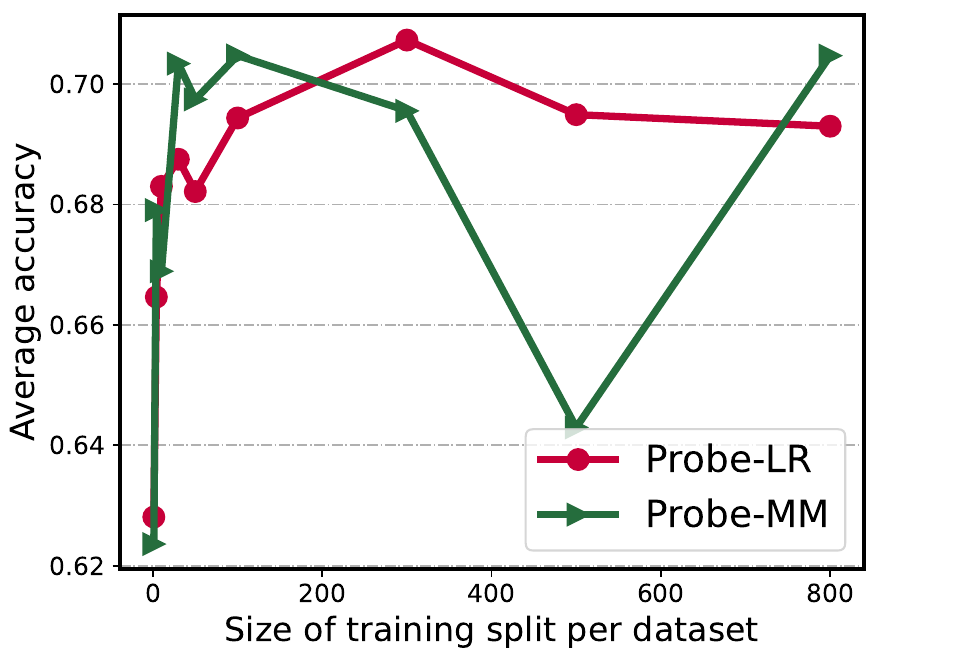} }

\vspace{-5pt}
	\caption{The analysis experiment results of training on attention head and layer activations, scaling number of training tasks, and varying training split size per task.}
	\label{fig-ablation2} 
  \vspace{-15pt}
\end{figure*}

\subsection{The Same Hyperplane in the Fine-tuned Models}

We conduct cross-task experiments to explore the generalization of our truthfulness hyperplane in the fine-tuned LLaMA2-7b-chat model (the FT model in Table~\ref{tab:main-result}). We directly evaluate our previous hyperplane trained from LLaMA2-7b-chat (without our fine-tuning) to classify hidden states from the fine-tuned LLaMA2-7b-chat model. Although the hyperplane is not trained on the finetuned model’s hidden states directly, we observe surprisingly higher accuracy when using it for the fine-tuned model than for the original LLaMA2-7b-chat that the hyperplane is trained on in Table~\ref{tab:hyperplane-gen}. The same hyperplane can generalize to the fine-tuned model and improve accuracy by 5 points, approaching the full-tuning accuracy. This suggests that after fine-tuning, the model has better truthfulness awareness and its inner hidden states are more linear-separable in terms of truthfulness.

\begin{table}[t]
    \centering
    \small
    \begin{tabular}{l  c}
        \toprule
        \textbf{} & \textbf{Cross-task Acc} \\
        \midrule
Fine-tuned Model	& 77.67 \\
LLaMA2-7b-chat Probe-LR &	69.24 \\
LLaMA2-7b-chat Probe-MM	& 70.47 \\
Fine-tuned Model Probe-LR &	75.16 \\
Fine-tuned Model Probe-MM &	74.46 \\
        \bottomrule
    \end{tabular}
    \vspace{-5pt}
    \caption{  The results of cross-task experiments using previous hyperplane evaluation on Fine-tuned Model and LLaMA2-7b-chat model.
    }
    \vspace{-5pt}
    \label{tab:hyperplane-gen}
\end{table}

\subsection{Analysis}
\label{analysis}
In this section, we perform a series of analysis and ablation experiments to justify our probe designs and gain deeper insights about the approach.
\paragraph{Which representation is better? Attention Heads or Layer Activations?}
\label{head-vs-layer}
In \textsection\ref{sec:probe}, we discussed the choice of input representation as part of the probe design and chose to use the attention heads in our main experiments.
Here we perform ablation on this design, comparing attention head and layer activations which are outputs after residual connections of the transformer layer.
Concretely, we train LR and MM probes using different numbers of training datasets on attention head outputs and layer residual activations respectively, conducting the cross-task experiments.
In Figure~\ref{fig-atten-layer} we show that
probes based on attention head outputs consistently outperform those trained on layer residual activations at least 3 points.
More setup details can be seen in Appendix~\ref{sec:appendix-preliminary}. As a result, we utilize the attention head output representations for training probes in our paper.

\paragraph{Effect of Number of Training Tasks:}

In light of the observed benefits of training on diverse datasets, a critical ablation study focuses on the impact of the quantity of training datasets on model performance. To investigate this, we incrementally increase the number of training tasks up to 14 (all training tasks),  with a corresponding increase in the number of datasets up to 41, conducting cross-task experiments of training on these incremented tasks. 
Our findings, illustrated in Figure~\ref{fig-scaling}, demonstrate a clear trend: as the number of training tasks increases, there is a general corresponding enhancement in average accuracy. This trend further indicates that
training on more diverse datasets helps to learn a more universal truthfulness hyperplane.
The Finetuning (FT) approach underperforms in comparison to the Probe method, when using one training task. This aligns with the observations reported by \citet{clymer2023generalization}. However, our study reveals a shift when the diversity of training datasets is expanded: the generalization performance of the FT method significantly outstrips that of the Probe method.

\vspace{-5pt}
\paragraph{Effect of Training Split Size for each Training Dataset:}

To explore the influence of sampled data volume for each dataset, we manipulate the training split size for each dataset and examine its effect on performance. The results are visualized in Figure~\ref{fig-scal-num}. Surprisingly, the results indicate that training even with as few as 10 data points per dataset, the performance is comparable to that of using 800 samples per dataset. This finding could be attributed to the probes' linear nature, making it not rely on extensive training data but only minimal data. These results are consistent with previous studies by \citet{li2023inference} and \citet{zou2023representation}, highlighting the effectiveness of training probes with limited data.

\section{Related Works}
Our work is related to a series of works trying to identify the truthfulness hyperplane inside LLMs. The existence of the universal truthfulness hyperplane is the foundation when considering truthfulness as an attribute for probing. Without such a hyperplane, it implies that all efforts in truthfulness probing~\citep{burns2022discovering,azaria2023internal,zou2023representation,marks2023geometry,li2023inference,chen2023truth} might merely be overfitting to spurious features of the task, rather than capturing genuine truthfulness.
Based upon such insights, several studies have also explored \textbf{interventions} to enhance model truthfulness by utilizing the vectors identified through probes~\citep{li2023inference,chen2023truth,zou2023representation}. Generally, utilizing the learned truthful vector, they edit the representation space directly~\citep{li2023inference,chen2023truth} or optimize the representation space towards more truthful states~\citep{zou2023representation}.

\section{Conclusion}
In this paper, we examine whether a universal truthfulness hyperplane exists inside the model, through designing and training a probe on diverse datasets. Our approach greatly improves existing results and conveys positive signals on the existence of such a universal truthfulness hyperplane.
\section*{Limitations}
\label{sec:significance}

First, there are several other methods to probe the language model's knowledge or hallucination, such as CCS \citep{burns2022discovering} and LAT \citep{zou2023representation}. In our paper, we only consider the commonly used supervised probing methods: logistic regression and mass mean. Further work can explore other methods. 
Second,  although we strive to include a wide range of diverse datasets, there is still a gap between our curated datasets and real-world data on truthfulness.
Third, we leave the intervention work as future research to verify whether the identified vector is causally related to model behavior.
Fourth,  although we are talking about truthfulness, the absolute detection accuracy is restricted by the knowledge of the model. The separation of correct and incorrect data within hidden representations is contingent upon the model's understanding. Consequently, our curated datasets may include noise stemming from the divergence between the model's knowledge and real-world knowledge, or from instances that exceed the model's knowledge boundaries.  We hypothesis that, in most cases, the knowledge of models aligns with the knowledge in data so that the Probe trained on our data can well discern the truthful or untruthful belief of the model.
Lastly, our experiments are limited to 7B and 13B size models, which demonstrate that stronger models exhibit a better truthfulness hyperplane. Future work can investigate whether the hidden states of even larger models, such as 70B models, are more linearly separable on truthfulness.




\bibliography{latex/custom}

\begin{thebibliography}{69}
\providecommand{\natexlab}[1]{#1}

\bibitem[{Alain and Bengio(2017)}]{alain2016understanding}
Guillaume Alain and Yoshua Bengio. 2017.
\newblock \href {https://openreview.net/forum?id=ryF7rTqgl} {Understanding intermediate layers using linear classifier probes}.

\bibitem[{Azaria and Mitchell(2023)}]{azaria2023internal}
Amos Azaria and Tom Mitchell. 2023.
\newblock \href {https://openreview.net/forum?id=y2V6YgLaW7} {The internal state of an {LLM} knows when it's lying}.
\newblock In \emph{The 2023 Conference on Empirical Methods in Natural Language Processing}.

\bibitem[{Belinkov(2022)}]{belinkov-2022-probing}
Yonatan Belinkov. 2022.
\newblock \href {https://doi.org/10.1162/coli_a_00422} {Probing classifiers: Promises, shortcomings, and advances}.
\newblock \emph{Computational Linguistics}, 48(1):207--219.

\bibitem[{Bisk et~al.(2020)Bisk, Zellers, Bras, Gao, and Choi}]{Bisk2020}
Yonatan Bisk, Rowan Zellers, Ronan~Le Bras, Jianfeng Gao, and Yejin Choi. 2020.
\newblock Piqa: Reasoning about physical commonsense in natural language.
\newblock In \emph{Thirty-Fourth AAAI Conference on Artificial Intelligence}.

\bibitem[{Brown et~al.(2020)Brown, Mann, Ryder, Subbiah, Kaplan, Dhariwal, Neelakantan, Shyam, Sastry, Askell et~al.}]{brown2020language}
Tom Brown, Benjamin Mann, Nick Ryder, Melanie Subbiah, Jared~D Kaplan, Prafulla Dhariwal, Arvind Neelakantan, Pranav Shyam, Girish Sastry, Amanda Askell, et~al. 2020.
\newblock Language models are few-shot learners.
\newblock \emph{Advances in neural information processing systems}, 33:1877--1901.

\bibitem[{Burns et~al.(2023)Burns, Ye, Klein, and Steinhardt}]{burns2022discovering}
Collin Burns, Haotian Ye, Dan Klein, and Jacob Steinhardt. 2023.
\newblock \href {https://openreview.net/forum?id=ETKGuby0hcs} {Discovering latent knowledge in language models without supervision}.
\newblock In \emph{The Eleventh International Conference on Learning Representations}.

\bibitem[{Campbell et~al.(2023)Campbell, Ren, and Guo}]{campbell2023localizing}
James Campbell, Richard Ren, and Phillip Guo. 2023.
\newblock Localizing lying in llama: Understanding instructed dishonesty on true-false questions through prompting, probing, and patching.
\newblock \emph{arXiv preprint arXiv:2311.15131}.

\bibitem[{CH-Wang et~al.(2023)CH-Wang, Van~Durme, Eisner, and Kedzie}]{ch2023androids}
Sky CH-Wang, Benjamin Van~Durme, Jason Eisner, and Chris Kedzie. 2023.
\newblock Do androids know they're only dreaming of electric sheep?
\newblock \emph{arXiv preprint arXiv:2312.17249}.

\bibitem[{Chen et~al.(2023)Chen, Sun, Jiao, Lian, Kang, Wang, and Xu}]{chen2023truth}
Zhongzhi Chen, Xingwu Sun, Xianfeng Jiao, Fengzong Lian, Zhanhui Kang, Di~Wang, and Cheng-Zhong Xu. 2023.
\newblock Truth forest: Toward multi-scale truthfulness in large language models through intervention without tuning.
\newblock \emph{arXiv preprint arXiv:2312.17484}.

\bibitem[{Chung et~al.(2022)Chung, Hou, Longpre, Zoph, Tay, Fedus, Li, Wang, Dehghani, Brahma et~al.}]{chung2022scaling}
Hyung~Won Chung, Le~Hou, Shayne Longpre, Barret Zoph, Yi~Tay, William Fedus, Yunxuan Li, Xuezhi Wang, Mostafa Dehghani, Siddhartha Brahma, et~al. 2022.
\newblock Scaling instruction-finetuned language models.
\newblock \emph{arXiv preprint arXiv:2210.11416}.

\bibitem[{Clark et~al.(2019)Clark, Lee, Chang, Kwiatkowski, Collins, and Toutanova}]{clark2019boolq}
Christopher Clark, Kenton Lee, Ming-Wei Chang, Tom Kwiatkowski, Michael Collins, and Kristina Toutanova. 2019.
\newblock \href {https://doi.org/10.18653/v1/N19-1300} {{B}ool{Q}: Exploring the surprising difficulty of natural yes/no questions}.
\newblock In \emph{Proceedings of the 2019 Conference of the North {A}merican Chapter of the Association for Computational Linguistics: Human Language Technologies, Volume 1 (Long and Short Papers)}, pages 2924--2936, Minneapolis, Minnesota. Association for Computational Linguistics.

\bibitem[{Clark et~al.(2018)Clark, Cowhey, Etzioni, Khot, Sabharwal, Schoenick, and Tafjord}]{allenai:arc}
Peter Clark, Isaac Cowhey, Oren Etzioni, Tushar Khot, Ashish Sabharwal, Carissa Schoenick, and Oyvind Tafjord. 2018.
\newblock Think you have solved question answering? try arc, the ai2 reasoning challenge.
\newblock \emph{arXiv:1803.05457v1}.

\bibitem[{Clymer et~al.(2023)Clymer, Baker, Subramani, and Wang}]{clymer2023generalization}
Joshua Clymer, Garrett Baker, Rohan Subramani, and Sam Wang. 2023.
\newblock Generalization analogies (genies): A testbed for generalizing ai oversight to hard-to-measure domains.
\newblock \emph{arXiv preprint arXiv:2311.07723}.

\bibitem[{Dalvi et~al.(2019)Dalvi, Durrani, Sajjad, Belinkov, Bau, and Glass}]{dalvi2019one}
Fahim Dalvi, Nadir Durrani, Hassan Sajjad, Yonatan Belinkov, Anthony Bau, and James Glass. 2019.
\newblock What is one grain of sand in the desert? analyzing individual neurons in deep nlp models.
\newblock In \emph{Proceedings of the AAAI Conference on Artificial Intelligence}, volume~33, pages 6309--6317.

\bibitem[{Du{\v{s}}ek et~al.(2020)Du{\v{s}}ek, Novikova, and Rieser}]{duvsek2020evaluating}
Ond{\v{r}}ej Du{\v{s}}ek, Jekaterina Novikova, and Verena Rieser. 2020.
\newblock Evaluating the state-of-the-art of end-to-end natural language generation: The e2e nlg challenge.
\newblock \emph{Computer Speech \& Language}, 59:123--156.

\bibitem[{Gardent et~al.(2017)Gardent, Shimorina, Narayan, and Perez-Beltrachini}]{gardent2017creating}
Claire Gardent, Anastasia Shimorina, Shashi Narayan, and Laura Perez-Beltrachini. 2017.
\newblock Creating training corpora for nlg micro-planning.
\newblock In \emph{55th annual meeting of the Association for Computational Linguistics (ACL)}.

\bibitem[{Geva et~al.(2021)Geva, Khashabi, Segal, Khot, Roth, and Berant}]{strategyqa}
Mor Geva, Daniel Khashabi, Elad Segal, Tushar Khot, Dan Roth, and Jonathan Berant. 2021.
\newblock \href {https://doi.org/10.1162/tacl_a_00370} {Did aristotle use a laptop? a question answering benchmark with implicit reasoning strategies}.
\newblock \emph{Transactions of the Association for Computational Linguistics}, 9:346--361.

\bibitem[{Gurnee et~al.(2023)Gurnee, Nanda, Pauly, Harvey, Troitskii, and Bertsimas}]{sparse}
Wes Gurnee, Neel Nanda, Matthew Pauly, Katherine Harvey, Dmitrii Troitskii, and Dimitris Bertsimas. 2023.
\newblock \href {https://openreview.net/forum?id=JYs1R9IMJr} {Finding neurons in a haystack: Case studies with sparse probing}.
\newblock \emph{Transactions on Machine Learning Research}.

\bibitem[{Hermann et~al.(2015)Hermann, Kocisky, Grefenstette, Espeholt, Kay, Suleyman, and Blunsom}]{hermann2015teaching}
Karl~Moritz Hermann, Tomas Kocisky, Edward Grefenstette, Lasse Espeholt, Will Kay, Mustafa Suleyman, and Phil Blunsom. 2015.
\newblock Teaching machines to read and comprehend.
\newblock \emph{Advances in neural information processing systems}, 28.

\bibitem[{Huang et~al.(2023)Huang, Yu, Ma, Zhong, Feng, Wang, Chen, Peng, Feng, Qin et~al.}]{huang2023survey}
Lei Huang, Weijiang Yu, Weitao Ma, Weihong Zhong, Zhangyin Feng, Haotian Wang, Qianglong Chen, Weihua Peng, Xiaocheng Feng, Bing Qin, et~al. 2023.
\newblock A survey on hallucination in large language models: Principles, taxonomy, challenges, and open questions.
\newblock \emph{arXiv preprint arXiv:2311.05232}.

\bibitem[{Jawahar et~al.(2019)Jawahar, Sagot, and Seddah}]{jawahar-etal-2019-bert}
Ganesh Jawahar, Beno{\^\i}t Sagot, and Djam{\'e} Seddah. 2019.
\newblock \href {https://doi.org/10.18653/v1/P19-1356} {What does {BERT} learn about the structure of language?}
\newblock In \emph{Proceedings of the 57th Annual Meeting of the Association for Computational Linguistics}, pages 3651--3657, Florence, Italy. Association for Computational Linguistics.

\bibitem[{Ji et~al.(2023)Ji, Lee, Frieske, Yu, Su, Xu, Ishii, Bang, Madotto, and Fung}]{ji2023survey}
Ziwei Ji, Nayeon Lee, Rita Frieske, Tiezheng Yu, Dan Su, Yan Xu, Etsuko Ishii, Ye~Jin Bang, Andrea Madotto, and Pascale Fung. 2023.
\newblock Survey of hallucination in natural language generation.
\newblock \emph{ACM Computing Surveys}, 55(12):1--38.

\bibitem[{Jiang et~al.(2023)Jiang, Sablayrolles, Mensch, Bamford, Chaplot, de~las Casas, Bressand, Lengyel, Lample, Saulnier, Lavaud, Lachaux, Stock, Scao, Lavril, Wang, Lacroix, and Sayed}]{jiang2023mistral}
Albert~Q. Jiang, Alexandre Sablayrolles, Arthur Mensch, Chris Bamford, Devendra~Singh Chaplot, Diego de~las Casas, Florian Bressand, Gianna Lengyel, Guillaume Lample, Lucile Saulnier, Lélio~Renard Lavaud, Marie-Anne Lachaux, Pierre Stock, Teven~Le Scao, Thibaut Lavril, Thomas Wang, Timothée Lacroix, and William~El Sayed. 2023.
\newblock \href {https://arxiv.org/abs/2310.06825} {Mistral 7b}.
\newblock \emph{Preprint}, arXiv:2310.06825.

\bibitem[{Jiang et~al.(2024)Jiang, Rajendran, Ravikumar, Aragam, and Veitch}]{jiang2024origins}
Yibo Jiang, Goutham Rajendran, Pradeep Ravikumar, Bryon Aragam, and Victor Veitch. 2024.
\newblock On the origins of linear representations in large language models.
\newblock \emph{arXiv preprint arXiv:2403.03867}.

\bibitem[{Joshi et~al.(2017)Joshi, Choi, Weld, and Zettlemoyer}]{2017arXivtriviaqa}
Mandar Joshi, Eunsol Choi, Daniel Weld, and Luke Zettlemoyer. 2017.
\newblock \href {https://doi.org/10.18653/v1/P17-1147} {{T}rivia{QA}: A large scale distantly supervised challenge dataset for reading comprehension}.
\newblock In \emph{Proceedings of the 55th Annual Meeting of the Association for Computational Linguistics (Volume 1: Long Papers)}, pages 1601--1611, Vancouver, Canada. Association for Computational Linguistics.

\bibitem[{Joshi et~al.(2023)Joshi, Rando, Saparov, Kim, and He}]{joshi2023personas}
Nitish Joshi, Javier Rando, Abulhair Saparov, Najoung Kim, and He~He. 2023.
\newblock Personas as a way to model truthfulness in language models.
\newblock \emph{arXiv preprint arXiv:2310.18168}.

\bibitem[{Kadavath et~al.(2022)Kadavath, Conerly, Askell, Henighan, Drain, Perez, Schiefer, Hatfield-Dodds, DasSarma, Tran-Johnson et~al.}]{kadavath2022language}
Saurav Kadavath, Tom Conerly, Amanda Askell, Tom Henighan, Dawn Drain, Ethan Perez, Nicholas Schiefer, Zac Hatfield-Dodds, Nova DasSarma, Eli Tran-Johnson, et~al. 2022.
\newblock Language models (mostly) know what they know.
\newblock \emph{arXiv preprint arXiv:2207.05221}.

\bibitem[{Khashabi et~al.(2018)Khashabi, Chaturvedi, Roth, Upadhyay, and Roth}]{khashabi2018looking}
Daniel Khashabi, Snigdha Chaturvedi, Michael Roth, Shyam Upadhyay, and Dan Roth. 2018.
\newblock Looking beyond the surface: A challenge set for reading comprehension over multiple sentences.
\newblock In \emph{Proceedings of the 2018 Conference of the North American Chapter of the Association for Computational Linguistics: Human Language Technologies, Volume 1 (Long Papers)}, pages 252--262.

\bibitem[{Kwiatkowski et~al.(2019)Kwiatkowski, Palomaki, Redfield, Collins, Parikh, Alberti, Epstein, Polosukhin, Devlin, Lee et~al.}]{kwiatkowski2019natural}
Tom Kwiatkowski, Jennimaria Palomaki, Olivia Redfield, Michael Collins, Ankur Parikh, Chris Alberti, Danielle Epstein, Illia Polosukhin, Jacob Devlin, Kenton Lee, et~al. 2019.
\newblock Natural questions: a benchmark for question answering research.
\newblock \emph{Transactions of the Association for Computational Linguistics}, 7:453--466.

\bibitem[{Lai et~al.(2017)Lai, Xie, Liu, Yang, and Hovy}]{lai2017race}
Guokun Lai, Qizhe Xie, Hanxiao Liu, Yiming Yang, and Eduard Hovy. 2017.
\newblock \href {https://doi.org/10.18653/v1/D17-1082} {{RACE}: Large-scale {R}e{A}ding comprehension dataset from examinations}.
\newblock In \emph{Proceedings of the 2017 Conference on Empirical Methods in Natural Language Processing}, pages 785--794, Copenhagen, Denmark. Association for Computational Linguistics.

\bibitem[{Levesque et~al.(2012)Levesque, Davis, and Morgenstern}]{levesque2012winograd}
Hector Levesque, Ernest Davis, and Leora Morgenstern. 2012.
\newblock The winograd schema challenge.
\newblock In \emph{Thirteenth international conference on the principles of knowledge representation and reasoning}.

\bibitem[{Li et~al.(2023{\natexlab{a}})Li, Cheng, Zhao, Nie, and Wen}]{li2023halueval}
Junyi Li, Xiaoxue Cheng, Xin Zhao, Jian-Yun Nie, and Ji-Rong Wen. 2023{\natexlab{a}}.
\newblock \href {https://openreview.net/forum?id=bxsrykzSnq} {Halueval: A large-scale hallucination evaluation benchmark for large language models}.
\newblock In \emph{The 2023 Conference on Empirical Methods in Natural Language Processing}.

\bibitem[{Li et~al.(2023{\natexlab{b}})Li, Patel, Vi{\'e}gas, Pfister, and Wattenberg}]{li2023inference}
Kenneth Li, Oam Patel, Fernanda Vi{\'e}gas, Hanspeter Pfister, and Martin Wattenberg. 2023{\natexlab{b}}.
\newblock \href {https://openreview.net/forum?id=aLLuYpn83y} {Inference-time intervention: Eliciting truthful answers from a language model}.
\newblock In \emph{Thirty-seventh Conference on Neural Information Processing Systems}.

\bibitem[{Lin et~al.(2022)Lin, Hilton, and Evans}]{lin2021truthfulqa}
Stephanie Lin, Jacob Hilton, and Owain Evans. 2022.
\newblock \href {https://doi.org/10.18653/v1/2022.acl-long.229} {{T}ruthful{QA}: Measuring how models mimic human falsehoods}.
\newblock In \emph{Proceedings of the 60th Annual Meeting of the Association for Computational Linguistics (Volume 1: Long Papers)}, pages 3214--3252, Dublin, Ireland. Association for Computational Linguistics.

\bibitem[{Maas et~al.(2011)Maas, Daly, Pham, Huang, Ng, and Potts}]{maas2011learning}
Andrew Maas, Raymond~E Daly, Peter~T Pham, Dan Huang, Andrew~Y Ng, and Christopher Potts. 2011.
\newblock Learning word vectors for sentiment analysis.
\newblock In \emph{Proceedings of the 49th annual meeting of the association for computational linguistics: Human language technologies}, pages 142--150.

\bibitem[{Marks and Tegmark(2023)}]{marks2023geometry}
Samuel Marks and Max Tegmark. 2023.
\newblock The geometry of truth: Emergent linear structure in large language model representations of true/false datasets.
\newblock \emph{arXiv preprint arXiv:2310.06824}.

\bibitem[{Meng et~al.(2022)Meng, Bau, Andonian, and Belinkov}]{meng2022locating}
Kevin Meng, David Bau, Alex Andonian, and Yonatan Belinkov. 2022.
\newblock Locating and editing factual associations in {GPT}.
\newblock \emph{Advances in Neural Information Processing Systems}, 36.

\bibitem[{Mihaylov et~al.(2018)Mihaylov, Clark, Khot, and Sabharwal}]{mihaylov2018can}
Todor Mihaylov, Peter Clark, Tushar Khot, and Ashish Sabharwal. 2018.
\newblock \href {https://doi.org/10.18653/v1/D18-1260} {Can a suit of armor conduct electricity? a new dataset for open book question answering}.
\newblock In \emph{Proceedings of the 2018 Conference on Empirical Methods in Natural Language Processing}, pages 2381--2391, Brussels, Belgium. Association for Computational Linguistics.

\bibitem[{Mostafazadeh et~al.(2017)Mostafazadeh, Roth, Louis, Chambers, and Allen}]{mostafazadeh2017lsdsem}
Nasrin Mostafazadeh, Michael Roth, Annie Louis, Nathanael Chambers, and James Allen. 2017.
\newblock Lsdsem 2017 shared task: The story cloze test.
\newblock In \emph{Proceedings of the 2nd Workshop on Linking Models of Lexical, Sentential and Discourse-level Semantics}, pages 46--51.

\bibitem[{Narayan et~al.(2018)Narayan, Cohen, and Lapata}]{Narayan2018DontGM}
Shashi Narayan, Shay~B. Cohen, and Mirella Lapata. 2018.
\newblock \href {https://doi.org/10.18653/v1/D18-1206} {Don{'}t give me the details, just the summary! topic-aware convolutional neural networks for extreme summarization}.
\newblock In \emph{Proceedings of the 2018 Conference on Empirical Methods in Natural Language Processing}, pages 1797--1807, Brussels, Belgium. Association for Computational Linguistics.

\bibitem[{Nie et~al.(2020)Nie, Williams, Dinan, Bansal, Weston, and Kiela}]{nie2019adversarial}
Yixin Nie, Adina Williams, Emily Dinan, Mohit Bansal, Jason Weston, and Douwe Kiela. 2020.
\newblock \href {https://doi.org/10.18653/v1/2020.acl-main.441} {Adversarial {NLI}: A new benchmark for natural language understanding}.
\newblock In \emph{Proceedings of the 58th Annual Meeting of the Association for Computational Linguistics}, pages 4885--4901, Online. Association for Computational Linguistics.

\bibitem[{Onoe et~al.(2021)Onoe, Zhang, Choi, and Durrett}]{onoe2021creak}
Yasumasa Onoe, Michael~JQ Zhang, Eunsol Choi, and Greg Durrett. 2021.
\newblock \href {https://openreview.net/forum?id=mbW_GT3ZN-} {{CREAK}: A dataset for commonsense reasoning over entity knowledge}.
\newblock In \emph{Thirty-fifth Conference on Neural Information Processing Systems Datasets and Benchmarks Track (Round 2)}.

\bibitem[{OpenAI(2023)}]{openai2023gpt}
OpenAI. 2023.
\newblock \href {https://arxiv.org/abs/2303.08774} {{GPT-4} technical report}.
\newblock \emph{arXiv preprint arXiv:2303.08774}.

\bibitem[{Rahman and Ng(2012)}]{rahman2012resolving}
Altaf Rahman and Vincent Ng. 2012.
\newblock Resolving complex cases of definite pronouns: the winograd schema challenge.
\newblock In \emph{Proceedings of the 2012 Joint Conference on Empirical Methods in Natural Language Processing and Computational Natural Language Learning}, pages 777--789.

\bibitem[{Rajpurkar et~al.(2016)Rajpurkar, Zhang, Lopyrev, and Liang}]{rajpurkar2016squad}
Pranav Rajpurkar, Jian Zhang, Konstantin Lopyrev, and Percy Liang. 2016.
\newblock \href {https://doi.org/10.18653/v1/D16-1264} {{SQ}u{AD}: 100,000+ questions for machine comprehension of text}.
\newblock In \emph{Proceedings of the 2016 Conference on Empirical Methods in Natural Language Processing}, pages 2383--2392, Austin, Texas. Association for Computational Linguistics.

\bibitem[{Roemmele et~al.(2011)Roemmele, Bejan, and Gordon}]{roemmele2011choice}
Melissa Roemmele, Cosmin~Adrian Bejan, and Andrew~S Gordon. 2011.
\newblock Choice of plausible alternatives: An evaluation of commonsense causal reasoning.
\newblock In \emph{2011 AAAI Spring Symposium Series}.

\bibitem[{Sakaguchi et~al.(2021)Sakaguchi, Bras, Bhagavatula, and Choi}]{sakaguchi2021winogrande}
Keisuke Sakaguchi, Ronan~Le Bras, Chandra Bhagavatula, and Yejin Choi. 2021.
\newblock Winogrande: An adversarial winograd schema challenge at scale.
\newblock \emph{Communications of the ACM}, 64(9):99--106.

\bibitem[{Sanh et~al.(2022)Sanh, Webson, Raffel, Bach, Sutawika, Alyafeai, Chaffin, Stiegler, Raja, Dey, Bari, Xu, Thakker, Sharma, Szczechla, Kim, Chhablani, Nayak, Datta, Chang, Jiang, Wang, Manica, Shen, Yong, Pandey, Bawden, Wang, Neeraj, Rozen, Sharma, Santilli, Fevry, Fries, Teehan, Scao, Biderman, Gao, Wolf, and Rush}]{sanh2021multitask}
Victor Sanh, Albert Webson, Colin Raffel, Stephen Bach, Lintang Sutawika, Zaid Alyafeai, Antoine Chaffin, Arnaud Stiegler, Arun Raja, Manan Dey, M~Saiful Bari, Canwen Xu, Urmish Thakker, Shanya~Sharma Sharma, Eliza Szczechla, Taewoon Kim, Gunjan Chhablani, Nihal Nayak, Debajyoti Datta, Jonathan Chang, Mike Tian-Jian Jiang, Han Wang, Matteo Manica, Sheng Shen, Zheng~Xin Yong, Harshit Pandey, Rachel Bawden, Thomas Wang, Trishala Neeraj, Jos Rozen, Abheesht Sharma, Andrea Santilli, Thibault Fevry, Jason~Alan Fries, Ryan Teehan, Teven~Le Scao, Stella Biderman, Leo Gao, Thomas Wolf, and Alexander~M Rush. 2022.
\newblock \href {https://openreview.net/forum?id=9Vrb9D0WI4} {Multitask prompted training enables zero-shot task generalization}.
\newblock In \emph{International Conference on Learning Representations}.

\bibitem[{Saxton et~al.(2019)Saxton, Grefenstette, Hill, and Kohli}]{saxton2019analysing}
David Saxton, Edward Grefenstette, Felix Hill, and Pushmeet Kohli. 2019.
\newblock \href {https://openreview.net/forum?id=H1gR5iR5FX} {Analysing mathematical reasoning abilities of neural models}.
\newblock In \emph{International Conference on Learning Representations}.

\bibitem[{See et~al.(2017)See, Liu, and Manning}]{see2017get}
Abigail See, Peter~J. Liu, and Christopher~D. Manning. 2017.
\newblock \href {https://doi.org/10.18653/v1/P17-1099} {Get to the point: Summarization with pointer-generator networks}.
\newblock In \emph{Proceedings of the 55th Annual Meeting of the Association for Computational Linguistics (Volume 1: Long Papers)}, pages 1073--1083, Vancouver, Canada. Association for Computational Linguistics.

\bibitem[{Srivastava et~al.(2023)Srivastava, Rastogi, Rao, Shoeb, Abid, Fisch, Brown, Santoro, Gupta, Garriga-Alonso et~al.}]{srivastava2022beyond}
Aarohi Srivastava, Abhinav Rastogi, Abhishek Rao, Abu Awal~Md Shoeb, Abubakar Abid, Adam Fisch, Adam~R Brown, Adam Santoro, Aditya Gupta, Adri{\`a} Garriga-Alonso, et~al. 2023.
\newblock \href {https://openreview.net/forum?id=uyTL5Bvosj} {Beyond the imitation game: Quantifying and extrapolating the capabilities of language models}.
\newblock \emph{Transactions on Machine Learning Research}.

\bibitem[{Sun et~al.(2019)Sun, Yu, Chen, Yu, Choi, and Cardie}]{sun2019dream}
Kai Sun, Dian Yu, Jianshu Chen, Dong Yu, Yejin Choi, and Claire Cardie. 2019.
\newblock Dream: A challenge data set and models for dialogue-based reading comprehension.
\newblock \emph{Transactions of the Association for Computational Linguistics}, 7:217--231.

\bibitem[{Talmor et~al.(2019)Talmor, Herzig, Lourie, and Berant}]{talmor2018commonsenseqa}
Alon Talmor, Jonathan Herzig, Nicholas Lourie, and Jonathan Berant. 2019.
\newblock \href {https://doi.org/10.18653/v1/N19-1421} {{C}ommonsense{QA}: A question answering challenge targeting commonsense knowledge}.
\newblock In \emph{Proceedings of the 2019 Conference of the North {A}merican Chapter of the Association for Computational Linguistics: Human Language Technologies, Volume 1 (Long and Short Papers)}, pages 4149--4158, Minneapolis, Minnesota. Association for Computational Linguistics.

\bibitem[{Tenney et~al.(2019)Tenney, Xia, Chen, Wang, Poliak, McCoy, Kim, Durme, Bowman, Das, and Pavlick}]{tenney2018what}
Ian Tenney, Patrick Xia, Berlin Chen, Alex Wang, Adam Poliak, R~Thomas McCoy, Najoung Kim, Benjamin~Van Durme, Sam Bowman, Dipanjan Das, and Ellie Pavlick. 2019.
\newblock \href {https://openreview.net/forum?id=SJzSgnRcKX} {What do you learn from context? probing for sentence structure in contextualized word representations}.
\newblock In \emph{International Conference on Learning Representations}.

\bibitem[{Tigges et~al.(2024)Tigges, Hollinsworth, Nanda, and Geiger}]{tigges2024language}
Curt Tigges, Oskar~John Hollinsworth, Neel Nanda, and Atticus Geiger. 2024.
\newblock \href {https://openreview.net/forum?id=iGDWZFc7Ya} {Language models linearly represent sentiment}.

\bibitem[{Touvron et~al.(2023{\natexlab{a}})Touvron, Lavril, Izacard, Martinet, Lachaux, Lacroix, Rozi{\`e}re, Goyal, Hambro, Azhar et~al.}]{llama}
Hugo Touvron, Thibaut Lavril, Gautier Izacard, Xavier Martinet, Marie-Anne Lachaux, Timoth{\'e}e Lacroix, Baptiste Rozi{\`e}re, Naman Goyal, Eric Hambro, Faisal Azhar, et~al. 2023{\natexlab{a}}.
\newblock Llama: Open and efficient foundation language models.
\newblock \emph{arXiv preprint arXiv:2302.13971}.

\bibitem[{Touvron et~al.(2023{\natexlab{b}})Touvron, Martin, Stone, Albert, Almahairi, Babaei, Bashlykov, Batra, Bhargava, Bhosale et~al.}]{llama2}
Hugo Touvron, Louis Martin, Kevin Stone, Peter Albert, Amjad Almahairi, Yasmine Babaei, Nikolay Bashlykov, Soumya Batra, Prajjwal Bhargava, Shruti Bhosale, et~al. 2023{\natexlab{b}}.
\newblock Llama 2: Open foundation and fine-tuned chat models.
\newblock \emph{arXiv preprint arXiv:2307.09288}.

\bibitem[{Vaswani et~al.(2017)Vaswani, Shazeer, Parmar, Uszkoreit, Jones, Gomez, Kaiser, and Polosukhin}]{vaswani2017attention}
Ashish Vaswani, Noam Shazeer, Niki Parmar, Jakob Uszkoreit, Llion Jones, Aidan~N Gomez, {\L}ukasz Kaiser, and Illia Polosukhin. 2017.
\newblock Attention is all you need.
\newblock \emph{Advances in neural information processing systems}, 30.

\bibitem[{Wang et~al.(2019)Wang, Singh, Michael, Hill, Levy, and Bowman}]{wang2018glue}
Alex Wang, Amanpreet Singh, Julian Michael, Felix Hill, Omer Levy, and Samuel~R. Bowman. 2019.
\newblock \href {https://openreview.net/forum?id=rJ4km2R5t7} {{GLUE}: A multi-task benchmark and analysis platform for natural language understanding}.
\newblock In \emph{International Conference on Learning Representations}.

\bibitem[{Wang et~al.(2023)Wang, Kordi, Mishra, Liu, Smith, Khashabi, and Hajishirzi}]{wang-etal-2023-self-instruct}
Yizhong Wang, Yeganeh Kordi, Swaroop Mishra, Alisa Liu, Noah~A. Smith, Daniel Khashabi, and Hannaneh Hajishirzi. 2023.
\newblock \href {https://doi.org/10.18653/v1/2023.acl-long.754} {Self-instruct: Aligning language models with self-generated instructions}.
\newblock In \emph{Proceedings of the 61st Annual Meeting of the Association for Computational Linguistics (Volume 1: Long Papers)}, pages 13484--13508, Toronto, Canada. Association for Computational Linguistics.

\bibitem[{Wei et~al.(2022)Wei, Bosma, Zhao, Guu, Yu, Lester, Du, Dai, and Le}]{wei2021finetuned}
Jason Wei, Maarten Bosma, Vincent Zhao, Kelvin Guu, Adams~Wei Yu, Brian Lester, Nan Du, Andrew~M. Dai, and Quoc~V Le. 2022.
\newblock \href {https://openreview.net/forum?id=gEZrGCozdqR} {Finetuned language models are zero-shot learners}.
\newblock In \emph{International Conference on Learning Representations}.

\bibitem[{Welbl et~al.(2017)Welbl, Liu, and Gardner}]{welbl2017crowdsourcing}
Johannes Welbl, Nelson~F. Liu, and Matt Gardner. 2017.
\newblock \href {https://doi.org/10.18653/v1/W17-4413} {Crowdsourcing multiple choice science questions}.
\newblock In \emph{Proceedings of the 3rd Workshop on Noisy User-generated Text}, pages 94--106, Copenhagen, Denmark. Association for Computational Linguistics.

\bibitem[{Yang et~al.(2018)Yang, Qi, Zhang, Bengio, Cohen, Salakhutdinov, and Manning}]{hotpotqa}
Zhilin Yang, Peng Qi, Saizheng Zhang, Yoshua Bengio, William Cohen, Ruslan Salakhutdinov, and Christopher~D. Manning. 2018.
\newblock \href {https://doi.org/10.18653/v1/D18-1259} {{H}otpot{QA}: A dataset for diverse, explainable multi-hop question answering}.
\newblock In \emph{Proceedings of the 2018 Conference on Empirical Methods in Natural Language Processing}, pages 2369--2380, Brussels, Belgium. Association for Computational Linguistics.

\bibitem[{Zellers et~al.(2019)Zellers, Holtzman, Bisk, Farhadi, and Choi}]{zellers2019hellaswag}
Rowan Zellers, Ari Holtzman, Yonatan Bisk, Ali Farhadi, and Yejin Choi. 2019.
\newblock \href {https://doi.org/10.18653/v1/P19-1472} {{H}ella{S}wag: Can a machine really finish your sentence?}
\newblock In \emph{Proceedings of the 57th Annual Meeting of the Association for Computational Linguistics}, pages 4791--4800, Florence, Italy. Association for Computational Linguistics.

\bibitem[{Zhang et~al.(2018)Zhang, Liu, Liu, Gao, Duh, and Van~Durme}]{zhang2018record}
Sheng Zhang, Xiaodong Liu, Jingjing Liu, Jianfeng Gao, Kevin Duh, and Benjamin Van~Durme. 2018.
\newblock Record: Bridging the gap between human and machine commonsense reading comprehension.
\newblock \emph{arXiv preprint arXiv:1810.12885}.

\bibitem[{Zhang et~al.(2015)Zhang, Zhao, and LeCun}]{zhang2015character}
Xiang Zhang, Junbo Zhao, and Yann LeCun. 2015.
\newblock Character-level convolutional networks for text classification.
\newblock \emph{Advances in neural information processing systems}, 28.

\bibitem[{Zhang et~al.(2019)Zhang, Baldridge, and He}]{paws2019naacl}
Yuan Zhang, Jason Baldridge, and Luheng He. 2019.
\newblock {PAWS: Paraphrase Adversaries from Word Scrambling}.
\newblock In \emph{Proc. of NAACL}.

\bibitem[{Zhang et~al.(2023)Zhang, Li, Cui, Cai, Liu, Fu, Huang, Zhao, Zhang, Chen et~al.}]{zhang2023siren}
Yue Zhang, Yafu Li, Leyang Cui, Deng Cai, Lemao Liu, Tingchen Fu, Xinting Huang, Enbo Zhao, Yu~Zhang, Yulong Chen, et~al. 2023.
\newblock Siren's song in the ai ocean: a survey on hallucination in large language models.
\newblock \emph{arXiv preprint arXiv:2309.01219}.

\bibitem[{Zou et~al.(2023)Zou, Phan, Chen, Campbell, Guo, Ren, Pan, Yin, Mazeika, Dombrowski et~al.}]{zou2023representation}
Andy Zou, Long Phan, Sarah Chen, James Campbell, Phillip Guo, Richard Ren, Alexander Pan, Xuwang Yin, Mantas Mazeika, Ann-Kathrin Dombrowski, et~al. 2023.
\newblock Representation engineering: A top-down approach to ai transparency.
\newblock \emph{arXiv preprint arXiv:2310.01405}.

\end{thebibliography}

\appendix

\section{Data Curation}
\label{sec:appendix}
We categorize datasets into one of the following task categories. For each dataset, we select a single prompt template to construct the dataset to reduce complexity. We utilize a maximum of 5000 data points for the test set for each dataset (if a dataset contains fewer than 5000 data points, we include all of them). 
Details of the used prompt and how to construct the wrong data points can be found below. 
\subsection{Natural language Inference}

\paragraph{RTE}

\noindent RTE is a testing textual entailment dataset \citep{wang2018glue}. We use one prompt template from \citet{sanh2021multitask}:\newline

Question: [premise]

      Does this mean that [hypothesis] is true? A) yes or B) no.

Answer: [label].\newline

\noindent Here [label] can be ``yes'' or ``no''. By selecting the opposite label, we construct the wrong data points.
\newline

\paragraph{QNLI}

\noindent The QNLI (Question Natural Language Inference) dataset is a collection of question-answer pairs, where the task is to determine whether the answer to a question is entailed in a given sentence \citep{wang2018glue}. We use one prompt template from \citet{sanh2021multitask}:\newline

Can you answer the question [question] based only on the following:

      [sentence]

Answer: [label].\newline

\noindent Here [label] can be ``yes'' or ``no'' By selecting the opposite label, we construct the wrong data points.
\newline

\paragraph{ANLI}

\noindent ANLI \citep{nie2019adversarial} is a difficult and adversarial  NLI dataset. We use one prompt template from \citet{sanh2021multitask}:\newline

[premise] Using only the above description and what you know about the
      world, [hypothesis] is definitely correct, incorrect, or inconclusive? 
      
Answer: [label]. \newline

\noindent Here [label] can be ``Correct'', ``Inconclusive'', or ``Incorrect''. By randomly selecting the wrong label, we construct the wrong data points.
\newline
\subsection{Summarization}

\paragraph{CNN Daily Mail}

\noindent CNN Daily Mail is a news summarization task \citep{hermann2015teaching, see2017get}. Given an article, the task is to generate the summary. We construct this dataset using the following prompt:\newline

Consider the accuracy of the summary of the following article.

Article: [article]

Summary: [summary]\newline


We leverage gpt-4-1106-preview to generate wrong summaries for CNN DailyMail dataset using the following instruction in Table~\ref{tab:cnn-instruction}, which is adapted from~\citet{li2023halueval}.

\begin{table}[h]
\centering
\small
\begin{tabular}{p{0.9\columnwidth}}
\toprule
I want you act as a hallucination summary generator. Given a document and the right summary, your objective is to write a hallucinated summary that sounds plausible but is factually incorrect. You SHOULD write the hallucinated summary using the following method (each with some examples):\\
\\
You are trying to write a summary but there is a factual contradiction between the summary and the document.\\
\#Document\#: Christopher Huxtable, 34, from Swansea, had been missing since the collapse in February. His body was found on Wednesday and workers who carried out the search formed a guard of honour as it was driven from the site in the early hours of the morning. Ken Cresswell, 57, and John Shaw, 61, both from Rotherham, remain missing. The body of a fourth man, Michael Collings, 53, from Brotton, Teesside, was previously recovered from the site. Swansea East MP Carolyn Harris, who has been involved with the family since the incident, said they still did not know all the facts about the collapse. She said: "I feel very sad. My heart and my prayers go out to the family who have waited desperately for Christopher's body to be found. They can finally have closure, and say goodbye to him and grieve his loss. "But let's not forget that there's two other families who are still waiting for their loved ones to be returned." The building was due for demolition when it partially collapsed in February.\\
\#Right Summary\#: A body found in the ruins of a collapsed building at Didcot Power Station has been identified.\\
\#Hallucinated Summary\#: The body of a man whose body was found at the site of the Swansea Bay Power Station collapse has been removed from the site.\\
\\
You should try your best to make the summary become hallucinated. \#Hallucinated Summary\# can only have about 5 more words than \#Right Summary\#.\\
\\
\#Document\#: [document]\\
\#Right Summary\#: [summary]\\
\#Hallucinated Summary\#:\\
\bottomrule
\end{tabular}
\caption{Instructions used for CNN DailyMail and XSum.}
\label{tab:cnn-instruction}
\end{table}

\paragraph{XSum}
Xsum is a summarization task with more concise summary \citep{Narayan2018DontGM}. We also use gpt4-1106-preview to generate wrong summaries using the same instruction as CNN Daily Mail in Table~\ref{tab:cnn-instruction}.
\newline
\subsection{Sentiment Analysis}
\paragraph{IMDB}
IMDB is a sentiment analysis dataset from \citet{maas2011learning}. Given a movie review, the task is to determine the sentiment is positive or negative. We use one prompt template from \citet{sanh2021multitask}:\newline

[review]

Is this review positive or negative?

[label].\newline

\noindent Here [label] can be ``Positive'' or ``Negative''. By selecting the opposite label, we construct the wrong data points.\newline

\paragraph{Yelp Polarity}
Yelp is a sentiment dataset from \citet{zhang2015character}. Given a yelp review, the task is to determine whether the review is good or bad. We use one prompt template from \citet{sanh2021multitask}:\newline

Review:

[review]

Overall rating (Good or Bad):

[label].\newline

\noindent Here [label] can be ``Good'' or ``Bad''. By selecting the opposite label, we can construct the wrong data points.\newline
\subsection{Topic Classification}
\paragraph{AG News}
AG News is a topic classification dataset from \citet{zhang2015character}. Given a news article, the task is to determine the topic of the article. We use one prompt template from \citet{sanh2021multitask}:\newline

Question: [text]

Which of the following sections of a newspaper would this article likely appear in? ``World News'', ``Sports'', ``Business'', or ``Science and Technology''?

Answer: [label].\newline

\noindent By selecting wrong label, we construct the wrong data points.\newline

\paragraph{DBPedia}
DBpedia is a topic classification dataset constructed by picking 14 non-overlapping classes from DBpedia 2014 \citet{zhang2015character}. We use the prompt template in \citet{burns2022discovering}:\newline

Consider the following example:

[text]

Which is the topic of this example, [label1] or [label2]?

[label].\newline

\noindent Here [label] can choose from ``Company'', ``Educational Institution'', ``Artist'', ``Athlete'',
      ``Office Holder'', ``Mean Of Transportation'', ``Building'', ``Natural Place'',
      ``Village'', ``Animal'', ``Plant'', ``Album'', ``Film'', ``Written Work''. By choosing the wrong label from [label1] and [label2], we construct the wrong data points.
\subsection{Statement Fact Checking}
\paragraph{Counterfact}
Couterfact is a model editing dataset with a correct target and a wrong target for a fact knowledge sentence \citep{meng2022locating}. By selecting correct targets or wrong targets, we construct correct data points and wrong data points. We directly use the sentence without any prompt template.\newline

[statement]\newline

\paragraph{Creak}
Creak is a dataset for commonsense reasoning over entity knowledge with sentences labeled true or false \citep{onoe2021creak}. Same as Counterfact, we don't use any prompt template.\newline

[statement]\newline

\paragraph{SAPLMA}
SAPLMA is a  true-false dataset with statements covering the following topics: ``Cities'', ``Inventions'', ``Chemical Elements'', ``Animals'', ``Companies'', and ``Scientific Facts'' \citep{azaria2023internal}.  Same as Counterfact and Creak, we directly use the statements as data points.\newline

[statement]\newline

\subsection{Paraphrase Identification}
\paragraph{MRPC}
MRPC dataset is a collection of sentence pairs with binary labels indicating whether the pair is a true paraphrase or not \citep{wang2018glue}. We use one prompt template from  \citet{sanh2021multitask}:\newline

Question: I want to know whether the following two sentences mean the same thing.

      [sentence1]

      [sentence2]

      Do they?

Answer: [label].\newline

\noindent Here [label] can be ``Yes'' or ``No''. By selecting the opposite label, we construct the wrong data points.\newline

\paragraph{QQP}
\noindent QQP dataset is a dataset consisting of pairs of questions, which labeled as either ``duplicate'' or ``not duplicate'', indicating whether the two questions are semantically equivalent or not \citep{wang2018glue}. We use one prompt template from \citet{sanh2021multitask}:\newline

Are the questions [question1] and [question2] asking the same thing?

Answer: [label].\newline

\noindent Here [label] can be ``Yes'' or ``No''. By choosing the opposite label, we construct the wrong data points.\newline

\paragraph{PAWS}
\noindent PAWS dataset consists of sentence pairs annotated as either semantically equivalent (i.e., paraphrases) or non-equivalent \citep{paws2019naacl}. We use one prompt template from \citet{sanh2021multitask}:\newline

Sentence 1: [sentence1]

Sentence 2: [sentence2]

Question: Do Sentence 1 and Sentence 2 express the same meaning? Yes or No? 

Answer: [label].\newline

\noindent Here [label] can be ``Yes'' or ``No''. By choosing the opposite label, we construct the wrong data points.\newline

\subsection{Short Answer Close Book QA}
\paragraph{Natural Questions}
Here we use nq open dataset consisting of questions (from Google Search) and short answers \citep{kwiatkowski2019natural}. We use the following prompt:\newline

Question: [question]

Answer: [answer]\newline

We leverage gpt-4-1106-preview to generate false answers, using the following instruction in Table~\ref{tab:nq-instruction}:

\begin{table}[h]
\centering
\small
\begin{tabular}{p{0.9\columnwidth}}
\toprule
Given a question and correct answer, you are asked to generate a reasonable but false answer. Here are some examples.\\
\#Qusetion\#: where did they film hot tub time machine\\
\#Correct Answer\#: Fernie Alpine Resort\\
\#False Answer\#: Town of Hobbiton, New Zealand\\
\\
\#Qusetion\#: who does annie work for attack on titan\\
\#Correct Answer\#: Marley\\
\#False Answer\#: The Survey Corps\\
\\
Here is the question and its correct answer, you need to generate a reasonable but false answer.\\
\#Question\#: [question]\\
\#Correct Answer\#: [answer]\\
\#False Answer\#:\\
\bottomrule
\end{tabular}
\caption{Instructions used for Natural Questions}
\label{tab:nq-instruction}
\end{table}

\paragraph{Trivia QA}
Trivia QA is a reading comprehension dataset containing over 650K question-answer-evidence triples \citep{2017arXivtriviaqa}. We only retain questions and answers and use the same prompt as Natural Questions.\newline

Question: [question]

Answer: [answer]\newline

We leverage gpt-4-1106-preview to generate false answers, using the following instruction in Table~\ref{tab:triviaqa-instruction}.

\begin{table}[h]
\centering
\small
\begin{tabular}{p{0.9\columnwidth}}
\toprule
Given a question and correct answer, you are asked to generate a reasonable but false answer. Here are some examples.\\
\#Question\#: Wolf Mankowitz wrote the 1953 novel ‘A Kid For Two\ldots’ what?\\
\#Correct Answer\#: Farthings\\
\#False Answer\#: Kookaburras\\\\
\#Question\#: The 2013-4 MacRobertson Shield international competition, hosted in New Zealand, was in what sport?\\
\#Correct Answer\#: Croquet\\
\#False Answer\#: Curling\\\\
Here is the question and its correct answer, you need to generate a reasonable but false answer.\\
\#Question\#: [question]\\
\#Correct Answer\#: [answer]\\
\#False Answer\#: \\
\bottomrule
\end{tabular}
\caption{Instructions used for Trivia QA}
\label{tab:triviaqa-instruction}
\end{table}

\paragraph{SciQ}
The SciQ dataset contains crowdsourced science exam questions about Physics, Chemistry and Biology, among others  with 4 answer options each \citep{welbl2017crowdsourcing}. We select one answer for each data and use same prompt as Natural Questions.\newline

Question: [question]

Answer: [answer]\newline

\noindent By selecting the wrong answer, we construct the wrong data points.\newline

\subsection{Long Answer Close Book QA}
\paragraph{Natural Questions Long}
To increase the diversity and better test generalization, we use gpt-4-1106-preview to rewrite the short answer in Natural Questions into one sentence long answer. Still, we use the same prompt template as Natural Questions.\newline

Question: [question]

Answer: [answer]\newline

We leverage gpt-4-1106-preview to paraphrase the short answer into a long answer in Natural Questions dataset using the following instruction in Table~\ref{tab:nq-long-instruction}.

\begin{table}[h]
\centering
\small
\begin{tabular}{p{0.9\columnwidth}}
\toprule
You need to rewrite the following short answers into a longer, complete sentence as the answer, even if the answer is incorrect, do not change the meaning.\\
\#Qusetion\#: where did the allies go after north africa\\
\#Short Answer\#: France\\
\#Long Answer\#: After the successful North African campaign, the Allies proceeded to advance towards France as part of their strategic plan during World War II.\\
\\
\#Qusetion\#: how many seasons of the bastard executioner are there\\
\#Short Answer\#: three\\
\#Long Answer\#: The Bastard Executioner" consists of a total of three seasons.\\
\\
Here is the question and its short answer, you only need to generate a long answer. Remember don't change the meaning, even if the answer is incorrect.\\
\#Question\#: [question]\\
\#Short Answer\#: [answer]\\
\#Long Answer\#:\\
\bottomrule
\end{tabular}
\caption{Instructions used for Natural Questions Long}
\label{tab:nq-long-instruction}
\end{table}

\paragraph{Trivia QA Long}
We also rewrite the short answer into long answer in Trivia QA to construct Trivia QA Long. We use the same prompt:\newline

Question: [question]

Answer: [answer]\newline

We leverage gpt-4-1106-preview to paraphrase the short answer into a long answer in Trivia QA dataset using the following instruction in Table~\ref{tab:triviaqa-long-instruction}.

\begin{table}[h]
\centering
\small
\begin{tabular}{p{0.9\columnwidth}}
\toprule
You need to rewrite the following short answers into a longer, complete sentence as the answer, even if the answer is incorrect, do not change the meaning.\\
\#Qusetion\#: Wolf Mankowitz wrote the 1953 novel ‘A Kid For Two\ldots’ what?\\
\#Short Answer\#: Pennies\\
\#Long Answer\#: Wolf Mankowitz, a notable author, penned the 1953 novel titled "A Kid For Two Pennies," showcasing his literary prowess and storytelling abilities.\\
\\
\#Qusetion\#: Who is the patron saint of dancers?\\
\#Short Answer\#: St. Cecilia\\
\#Long Answer\#: St. Cecilia, a revered figure in religious history, holds the esteemed title of being the patron saint specifically designated to protect and guide dancers, bestowing upon them blessings and interceding on their behalf.\\
\\
Here is the question and its short answer, you only need to generate a long answer. Remember don't change the meaning, even if the answer is incorrect.\\
\#Question\#: [question]\\
\#Short Answer\#: [answer]\\
\#Long Answer\#:\\
\bottomrule
\end{tabular}
\caption{Instructions used for Trivia QA Long}
\label{tab:triviaqa-long-instruction}
\end{table}

\subsection{Reading Comprehension (QA)}
\paragraph{MultiRC}
MultiRC (Multi-Sentence Reading Comprehension) is a dataset of short paragraphs and multi-sentence questions with answers labeled true or false \citep{khashabi2018looking}. We use the following prompt:\newline

Exercise: read the text and answer the question.

Text: [passage]

Question: [question]

Answer: [answer]\newline

\noindent Since MultiRC already has labeled wrong answers, we construct the wrong data points using the wrong answers.\newline

\paragraph{SQuAD}
SQuAD is a reading comprehension dataset, consisting of questions on a set of Wikipedia articles, where the answer to every question is a segment of text, or span, from the corresponding reading passage, or the question might be unanswerable \citep{rajpurkar2016squad}. We use
one prompt template from \citet{sanh2021multitask}:\newline

Refer to the passage below and answer the following question:

      Passage: [context]

      Question: [question]

      Answer: [answer]\newline

We use gpt-4-1106-preview to generate false answers for SQuAD dataset using the instruction in Table~\ref{tab:squad-instruction}.

\begin{table}[h]
\centering
\small
\begin{tabular}{p{0.9\columnwidth}}
\toprule
Given a passage, a question and the right answer, your objective is to write a answer that sounds plausible (appears in the passage) but is incorrect. Here is an example.\\
\#Passage\#: Super Bowl 50 was an American football game to determine the champion of the National Football League (NFL) for the 2015 season. The American Football Conference (AFC) champion Denver Broncos defeated the National Football Conference (NFC) champion Carolina Panthers 24–10 to earn their third Super Bowl title. The game was played on February 7, 2016, at Levi's Stadium in the San Francisco Bay Area at Santa Clara, California. As this was the 50th Super Bowl, the league emphasized the "golden anniversary" with various gold-themed initiatives, as well as temporarily suspending the tradition of naming each Super Bowl game with Roman numerals (under which the game would have been known as "Super Bowl L"), so that the logo could prominently feature the Arabic numerals 50.\\
\#Question\#: Where did Super Bowl 50 take place?\\
\#Correct Answer\#: Santa Clara, California\\
\#False Answer\#: San Francisco, California\\
\\
\#Passage\#: Archaeological evidence shows that Homo erectus lived in the region now known as Myanmar as early as 400,000 years ago. The first evidence of Homo sapiens is dated to about 11,000 BC, in a Stone Age culture called the Anyathian with discoveries of stone tools in central Myanmar. Evidence of neolithic age domestication of plants and animals and the use of polished stone tools dating to sometime between 10,000 and 6,000 BC has been discovered in the form of cave paintings near the city of Taunggyi.\\
\#Question\#: When was the extinct species believed to have lived in Myanmar?\\
\#Correct Answer\#: 400,000 years ago\\
\#False Answer\#: 11,000 BC\\
\\
Here is the passage question and its correct answer, you need to generate a reasonable but false answer.\\
\#Passage\#: [passage]\\
\#Question\#: [question]\\
\#Correct Answer\#: [answer]\\
\#False Answer\#:\\
\bottomrule
\end{tabular}
\caption{Instructions used for SQuAD}
\label{tab:squad-instruction}
\end{table}

\subsection{Reading comprehension multi-choice}
\paragraph{BoolQ}
BoolQ is a question answering dataset for yes/no questions with passages \citep{clark2019boolq}. We use the following prompt:\newline

Passage: [passage]

After reading this passage, I have a question: [question]? True or False?

Answer: [answer].\newline

\noindent [answer] can be ``True'' or ``False''. By selecting the opposite answer, we construct the wrong data points.\newline

\paragraph{RACE}
RACE is a reading comprehension dataset with passages, questions and four choices collected from English examinations in China, which are designed for middle school and high school students \citep{lai2017race}. We use one prompt template in \citet{sanh2021multitask}.\newline

I'm taking a test and have to guess the right answer to the question
      after the article.

      Article: [article]

      Question: [question]

      Options: A: [options.0]

      B: [options.1]

      C: [options.2]

      D: [options.3]

      Answer: [answer].\newline

\noindent [answer] can be ``A'', ``B'', ``C'' or ``D''. By selecting the wrong answer, we construct the wrong data points.\newline

\paragraph{DREAM}
DREAM is a multiple-choice Dialogue-based Reading comprehension examination dataset. In contrast to existing reading comprehension datasets \citep{sun2019dream}. We use one prompt template from \citet{sanh2021multitask}:\newline

Dialogue:

[dialogue]

Question: [question]

- choices[0]

- choices[1]

- choices[2]

Answer: [answer]\newline

\noindent [answer] is selected from three choices. By selecting wrong choices, we construct the wrong data points.\newline

\subsection{Sentence Completion}
\paragraph{CoPA}
CoPA is a causal reasoning task to determine either the cause or the effect of a given premise \citep{roemmele2011choice}. We use one prompt template in \citet{sanh2021multitask}:\newline

Exercise: choose the most plausible alternative.

      [ premise ] 
      
      \{ if [question] == ``cause'' \} because... \{ else \} so... \{ endif
      \}

      - [choice1]

      - [choice2] 
      
Answer: [answer]\newline

\noindent [answer] is selected from the two choices. By selecting the wrong choice, we construct the wrong data points.\newline

\paragraph{HellaSwag}
Hellaswag dataset is a benchmark dataset created for the task of commonsense reasoning and understanding, specifically for the task of predicting the correct continuation of a given sentence \citep{zellers2019hellaswag}. We use one prompt template from \citet{sanh2021multitask}:\newline

Complete the description with an appropriate ending:

      First, [sentence1] Then, [sentence2] ...

      (a) choices[0]

      (b) choices[1]

      (c) choices[2]

      (d) choices[3]

Answer: [answer]\newline

\noindent [answer] is selected from the four choices. By selecting the wrong choices randomly, we construct the wrong data points.\newline

\paragraph{Story Cloze}
Story Cloze is a commonsense reasoning dataset for evaluating the choosing the correct ending to a four-sentence story ability \citep{mostafazadeh2017lsdsem}. We use one prompt template from \citet{sanh2021multitask}:\newline

[sentence1] [sentence2] [sentence3] [sentence4]

      What is a possible continuation for the story given the following options ?
      
      - choices[0]

      - choices[1]

Answer: [answer]\newline

\noindent [answer] is selected from two choices. By selecting the wrong choices, we construct the wrong data points.\newline

\subsection{Close Book Multi-Choice QA}
\paragraph{CommonsenseQA}
CommonsenseQA is a multiple-choice question answering dataset that requires different types of commonsense knowledge to predict the correct answers \citep{talmor2018commonsenseqa}. We use one prompt template from \citet{sanh2021multitask}:\newline

Question: Given the following options, what do you think is the correct answer to the
      question below:
      
      [question]

      Options:

      - A: choices[0]

    - B: choices[1]

      - C: choices[2]

    - D: choices[3]
    
    - E: choices[4]
    
Answer: [answer].\newline

\noindent [answer] is selected from ``A'', ``B'', ``C'', ``D'', ``E''. By randomly selecting wrong answers, we construct the wrong data points.\newline

\paragraph{ARC}
ARC is a multi-choice QA dataset which requires knowledge and reasoning \citep{allenai:arc}. It includes challenge and easy parts. We use both parts.\newline
For arc easy part, we use one prompt template in \citet{sanh2021multitask}:\newline

[question]

      Options:

- choices[0]

- choices[1]

- choices[2]

- choices[3]

Answer: [answer]\newline

\noindent Here [answer] is selected from the two choices. By selecting wrong choices randomly, we construct the wrong data points.\newline
For arc challenge part, we also use one prompt template in \citet{sanh2021multitask}:\newline

Here's a problem to solve: [question]

Among the 4 following options, which is the correct answer?

      - A: choices[0]

    - B: choices[1]

      - C: choices[2]

    - D: choices[3]

Answer: [answer].\newline

\noindent Here [answer] is selected from ``A'', ``B'', ``C'', ``D''. We construct wrong data points by selecting wrong answer.\newline

\paragraph{PIQA}
PIQA is a dataset requiring physical commonsense reasoning. Given a question q and two possible solutions s1, s2, the task is to choose the most appropriate solution \citep{Bisk2020}. We use one prompt template in \citet{sanh2021multitask}:\newline

Solution 1: [sol1]

Solution 2: [sol2]

      Goal: [goal]

      Given the goal, what is the correct solution?

      Answer by copying the correct solution

Answer: [answer]\newline

\noindent Here [answer] is selected from two sol choices. By selecting  wrong choices, we construct wrong data points.\newline

\paragraph{OpenBookQA}
OpenBookQA contains questions that require reasoning and commonsense knowledge \citep{mihaylov2018can}. The task is to select correct answer from four choices for the given question. We use one prompt template in \citet{sanh2021multitask}:\newline

Question: [question]

      Choose an answer from this list:

      - choices[0]

      - choices[1]

    - choices[2]

      - choices[3]

Answer: [answer]\newline

\noindent Here [answer] is selected from the four choices. By selecting wrong choices, we construct wrong data points.\newline

\subsection{Structure To Text}
\paragraph{E2ENLG}
Here we use E2ENLG CLEAN dataset. The E2E NLG dataset is a dataset for the task of data-to-text natural language generation \citep{duvsek2020evaluating}. It consists of tables containing structured data, and corresponding human-written textual descriptions of that data. We use one prompt template in \citep{sanh2021multitask}:\newline

Combine all of the following data into a concise and grammatically correct
 text:

        key1: value1

      key2: value2

      ...
      
      Generated\_text: [human\_reference]\newline

\noindent Following the synthetic hallucinations method mentioned in \citet{ch2023androids}, for an example with $n$ attributes, we modify $k$ attributes (drawn uniformly from $[1, n-1]$) by replacing their values with other values that correspond to the same key. Using the resulting modified data and keeping [text] unchanged, we construct wrong data points.\newline

\paragraph{WEBNLG}
WebNLG dataset is mapping data to text, where the data is a set of triples extracted from DBpedia and the text is a verbalisation of these triples \citep{gardent2017creating}. We use one prompt template in \citet{sanh2021multitask}:\newline

Take the following triple set as part of a Data-to-Text task: [data]. Make a lexicalization of the triple set into plain text.
      
      Generated text: [text]\newline

\noindent We use gpt-3.5-turbo to modify the attributes and then generate new text using the instruction in Table~\ref{tab:webnlg-instruction}.

\begin{table}[h]
\centering
\small
\begin{tabular}{p{0.9\columnwidth}}
\toprule

Given the mtriple\_set data and its corresponding plain text, you are asked to modify some (but not all) of the feature information in the mtriple\_set and generate a new text based on the new mtriple\_set. Here are some examples.\\
\#mtriple\_set\#: [\\
"Pontiac\_Rageous | productionStartYear | 1997",\\
"Pontiac\_Rageous | assembly | Michigan"\\
]\\
\#text\#: The Pontiac Rageous was first produced in 1997 in Michigan.\\
\#new mtriple\_set\#: [\\
"Pontiac\_Rageous | productionStartYear | 1997",\\
"Pontiac\_Rageous | assembly | Ohio"\\
]\\
\#new text\#: The initial production of the Pontiac Rageous took place in 1997 in Ohio.\\
\\
\#mtriple\_set\#: [\\
"Acharya\_Institute\_of\_Technology | president | "B.M. Reddy"",\\
"Acharya\_Institute\_of\_Technology | city | Bangalore",\\
"Acharya\_Institute\_of\_Technology | established | 2000",\\
"Acharya\_Institute\_of\_Technology | country | "India"",\\
"Acharya\_Institute\_of\_Technology | state | Karnataka",\\
"Acharya\_Institute\_of\_Technology | numberOfPostgraduateStudents | 700",\\
"Acharya\_Institute\_of\_Technology | campus | "In Soldevanahalli, Acharya Dr. Sarvapalli Radhakrishnan Road, Hessarghatta Main Road, Bangalore – 560090.""\\
] \\
\#text\#: Acharya Institute of Technology (president B M Reddy) was established in 2000 and has 700 postgraduate students. The campus is located at Soldevanahalli, Acharya Dr. Sarvapalli Radhakrishnan Road, Hessarghatta Main Road, Bangalore – 560090, Karnataka, India.\\
\#new mtriple\_set\#: [\\
"Acharya\_Institute\_of\_Technology | president | Mr. B.G. Reddy",\\
"Acharya\_Institute\_of\_Technology | city | Mysore",\\
"Acharya\_Institute\_of\_Technology | established | 2000",\\
"Acharya\_Institute\_of\_Technology | country | India",\\
"Acharya\_Institute\_of\_Technology | state | Karnataka",\\
"Acharya\_Institute\_of\_Technology | numberOfPostgraduateStudents | 700",\\
"Acharya\_Institute\_of\_Technology | campus | In Soldevanahalli, Acharya Dr. Sarvapalli Radhakrishnan Road, Hessarghatta Main Road, Mysore – 560090."\\
]\\
\#new text\#: Acharya Institute of Technology, located in Mysore, Karnataka, India, was established in the year 2000. Under the leadership of President Mr. B.G. Reddy, the institute has grown to accommodate 700 postgraduate students. The campus is situated in Soldevanahalli, on Acharya Dr. Sarvapalli Radhakrishnan Road, Hessarghatta Main Road, Mysore – 560090.\\
\\
Here is the test.\\
\#mtriple\_set\#: [mtriple\_set]\\
\#text\#: [text]\\
\#new mtriple\_set\#:\\

\bottomrule
\end{tabular}
\caption{Instructions used for WEBNLG}
\label{tab:webnlg-instruction}
\end{table}

\subsection{Coreference}
\paragraph{Definite Pronoun Resolution}
Definite Pronoun Resolution (DPR) dataset is a collection of annotated sentences that are used to train and evaluate models for resolving definite pronouns in English text \citep{rahman2012resolving}. Given a pronoun, the task is to select the correct antecedent noun phrase that the pronoun refers to. We use the following prompt:\newline

Question: [sentence]

Who is [pronoun] referring to?

[candidate1] or [candidate2]

Answer: [answer].\newline

\noindent [answer] is selected from [candidate1] and [candidate2]. By selecting wrong candidates, we construct wrong data points.\newline

\paragraph{Winogrande}
Here we use Winograde xl version. Winogrande is a dataset to test a machine's ability to understand natural language in context and resolve ambiguities \citep{sakaguchi2021winogrande}. With binary options, the goal is to choose the right option for a given sentence. We use one prompt template in \citet{sanh2021multitask}:\newline

Question: [sentence] In the previous sentence, does \_ refer to [option1]
      or  [option2]? 

Answer: [answer].\newline

\noindent [answer] is selected from two options. By selecting wrong options, we construct wrong data points.\newline

\paragraph{WSC.Fixed}
WSC  Fixed dataset is a collection of pronoun resolution problems used for evaluating natural language understanding systems. The goal is to determine the correct referent for the pronoun in each sentence \citep{levesque2012winograd}. We use one prompt template in \citet{sanh2021multitask}:\newline

[text] In the previous sentence, does the pronoun ``[pronoun]'' refer to [noun]? Yes or no?

[answer].\newline

\noindent Here [answer] is ``Yes'' or ``No''. By selecting the opposite answer, we construct the wrong data points.\newline

\subsection{Reading Comprehension and Common Sense}
\paragraph{ReCoRD}
Reading Comprehension with Commonsense Reasoning Dataset (ReCoRD) is a large-scale reading comprehension dataset which requires commonsense reasoning. ReCoRD consists of queries automatically generated from CNN/Daily Mail news articles; the answer to each query is a text span from a summarizing passage of the corresponding news \citep{zhang2018record}. We use one prompt template in \citet{sanh2021multitask}:\newline

[passage]

[query] 

You should decide what ``@placeholder'' is referring to. Choose between:

-  choices[0]

-  choices[1]

...

Answer: [answer].\newline

\noindent Here [answer] is selected from choices. By selecting wrong choices, we construct wrong data points.\newline

\paragraph{CosmosQA}
CosmosQA is a  dataset of  problems that require commonsense-based reading comprehension, formulated as multiple-choice questions. It focuses on  people’s everyday narratives, asking questions concerning on the likely causes or effects of events that require reasoning beyond the exact text spans in the context. We use one prompt template in \citet{sanh2021multitask}:\newline

[context]

      According to the above context, choose the best option to answer the following question.

      Question: [question]

      Options:

-  choices[0]

-  choices[1]

...

      Answer: [answer]\newline

\noindent Here [answer] is selected from choices. By selecting wrong choices, we construct wrong data points.\newline

\subsection{Multi-step Reasoning QA}
\paragraph{HotpotQA}
HotpotQA is a question answering dataset where the questions require finding and reasoning over multiple supporting documents to answer \citep{hotpotqa}. We use the following prompt:\newline

Questino: [question]

Answer: [answer]\newline

\noindent We leverage gpt-4-1106-preview to generate false answers, using the following instruction in Table~\ref{tab:Hotpot QA-instruction}:

\begin{table}[t]
\centering
\small
\begin{tabular}{p{0.9\columnwidth}}
\toprule
Given a question and correct answer, you are asked to generate a reasonable but false answer. Here are some examples.\\
\#Qusetion\#: What nationality was James Henry Miller's wife?\\
\#Correct Answer\#: American\\
\#False Answer\#: British\\
\\
\#Qusetion\#: British band The Wanted's third album includes a song with a title about which Barbadian superstar?\\
\#Correct Answer\#: Rihanna\\
\#False Answer\#: Shakira\\
\\
Here is the question and its correct answer, you need to generate a reasonable but false answer.\\
\#Question\#: [question]\\
\#Correct Answer\#: [answer]\\
\#False Answer\#:\\
\bottomrule
\end{tabular}
\caption{Instructions used for Hotpot QA}
\label{tab:Hotpot QA-instruction}
\end{table}

\paragraph{Strategy QA}
StrategyQA is a question-answering benchmark focusing on open-domain questions where the required reasoning steps are implicit in the question and should be inferred using a strategy \citep{strategyqa}. We use the following prompt:\newline

Question: [question]

Answer: [answer].\newline

\noindent Here [answer] can be ``Yes'' or ``No''. By selecting the opposite answers, we construct the wrong data points.\newline

\subsection{Other}
\paragraph{Truthful QA}
TruthfulQA is a benchmark to measure whether a language model is truthful in generating answers to questions where questions are crafted so that some humans would answer falsely due to a false belief or misconception \citep{lin2021truthfulqa}. We use the following prompt:\newline

Question: [question]

Answer: [answer]\newline

\noindent By selecting false answers in the dataset, we construct the wrong data points.\newline

\paragraph{Arithmetic}
Arithmetic dataset is a QA dataset comprising straightforward questions involving addition, subtraction, multiplication, and division \citep{saxton2019analysing, brown2020language}. We use the dataset in \citet{srivastava2022beyond}. We use the following prompt:\newline

Question: [question]

Answer: [answer]\newline

\noindent We use the given wrong answer in the dataset when constructing the wrong data points.

\section{Ablation study on hyperparameter $num$}
\label{sec:appendix-hyperpara-num}
\begin{figure} [h]
  \centering
  \includegraphics[scale=0.45]{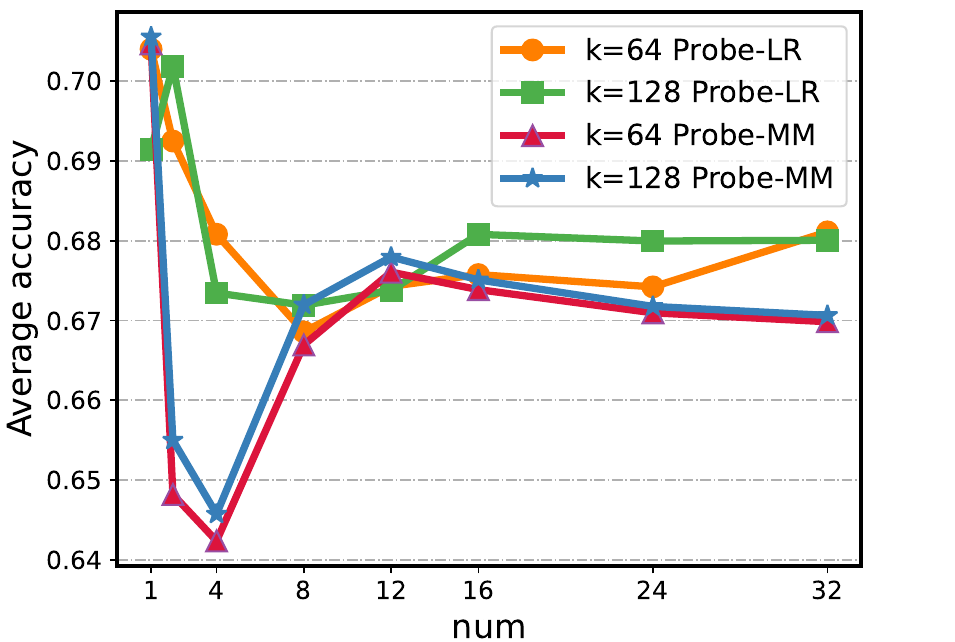}
  \caption{Ablation study of varying $num$ on cross-task test, where $k$ is the compression hyperparameter (128 represents all dimensions of the attention head output).}
  \label{fig-ablation-num}
\end{figure}

$num$ is the hyperparameter that determine the number of selected positions for each validation split.
Here, we conduct ablation studying on $num$. Varying the $num$, we train probes on all our curated training tasks, selecting $num$ positions for every validation split in training tasks and evaluate on the test tasks in Figure~\ref{fig-dataset}.
The results in Figure~\ref{fig-ablation-num} show that $num$ is 1 or 2 yields highest performance, while including more positions for every validation split even leads to a slight performance decline. Besides, increasing $num$ also leads to more memory and time cost.

\section{Sparisty}
\label{sec:appendix-sparsity}
\begin{figure*} [t]
	\centering
	\subfloat[Natural Questions]{
	\hspace{-0.3cm}	
  \includegraphics[scale=0.22]{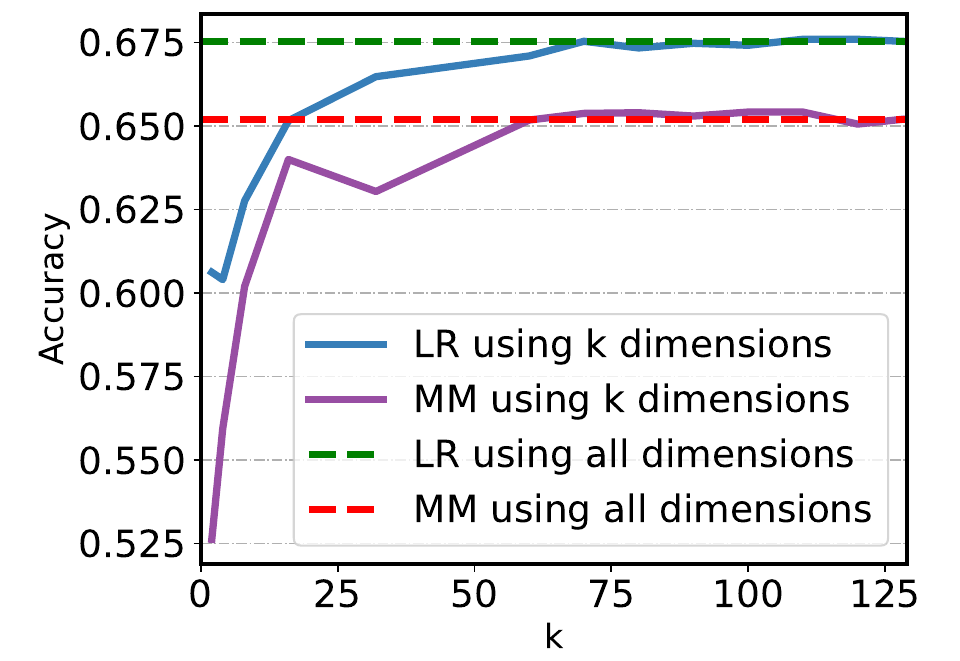}}
	\subfloat[TruthfulQA]{
		\includegraphics[scale=0.22]{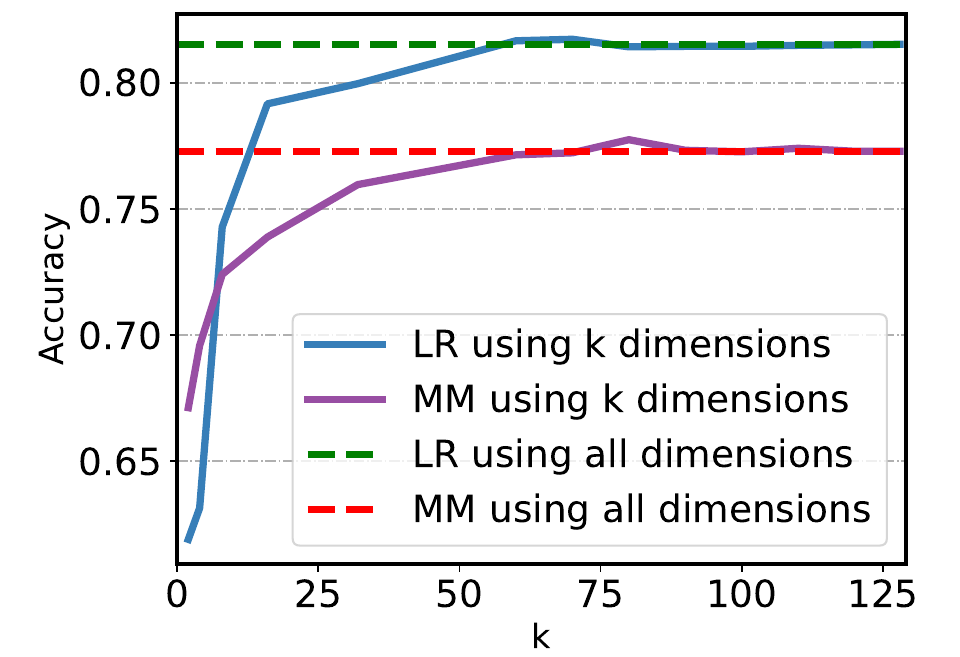} }
	\subfloat[Sciq]{
        \hspace{-0.3cm}
		\includegraphics[scale=0.22]{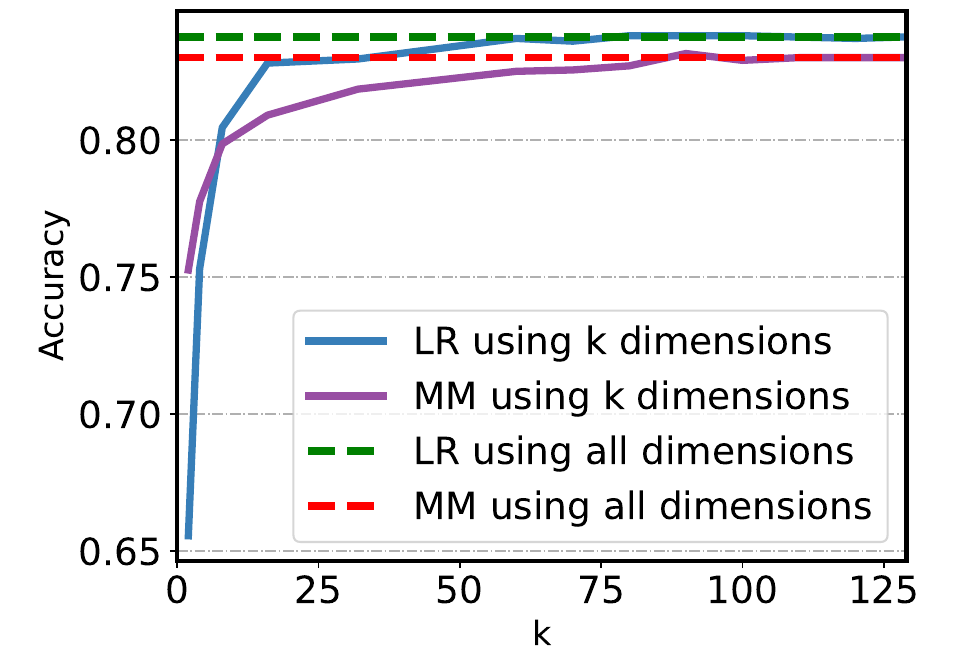} }
	\subfloat[Creak]{
        \hspace{-0.3cm}
		\includegraphics[scale=0.22]{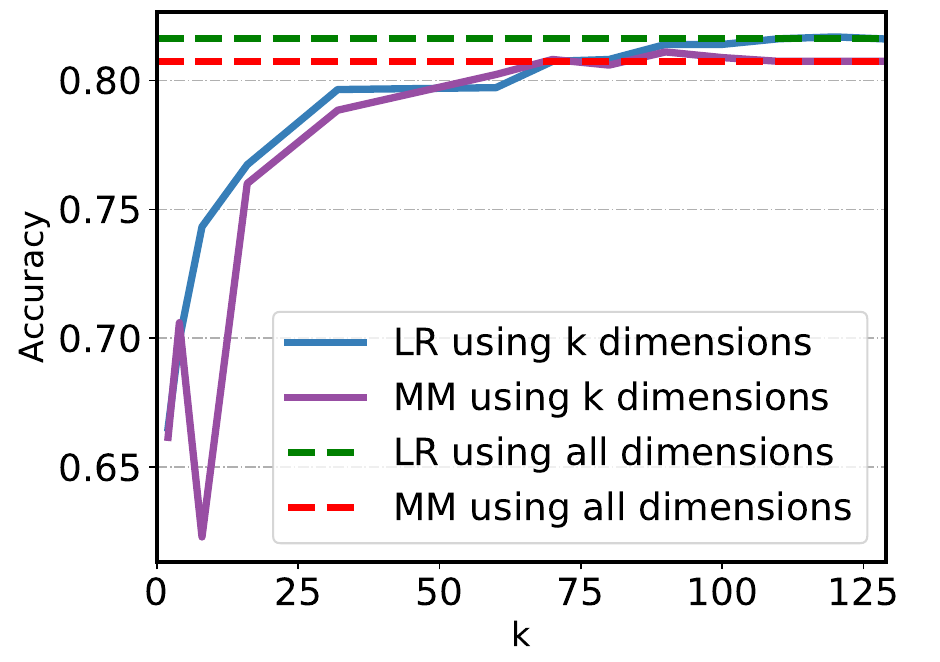} }
	\caption{Some other sparsity observations of attention head outputs on different tasks using the logistic regression (LR) and the mass mean (MM) probe. 
 }
	\label{fig-sparsity-show2} 
\end{figure*}

In this experiment, we study the sparsity by training probes on the training set of a single dataset and evaluating them on the corresponding test split. We train probes for every attention head output and then select the position with the highest accuracy to study the sparsity of the representation. Using the ranking method described in \textsection\ref{compression}, we first compress the full dimensions of the attention head output to varying $k$ dimensions. Then we retrain probes using the compressed representations and test the newly trained probes on the test split. Figure \ref{fig-sparsity-show2} displays more results.
Our results indicate that using half the dimensions of the attention head output is sufficient to achieve performance comparable to using the full dimensions. Therefore, we set the hyperparameter $k$ to be 64 or 128.

\begin{figure*} [t]
	\centering
	\subfloat[Natural Questions]{
	\hspace{-0.3cm}	
  \includegraphics[scale=0.22]{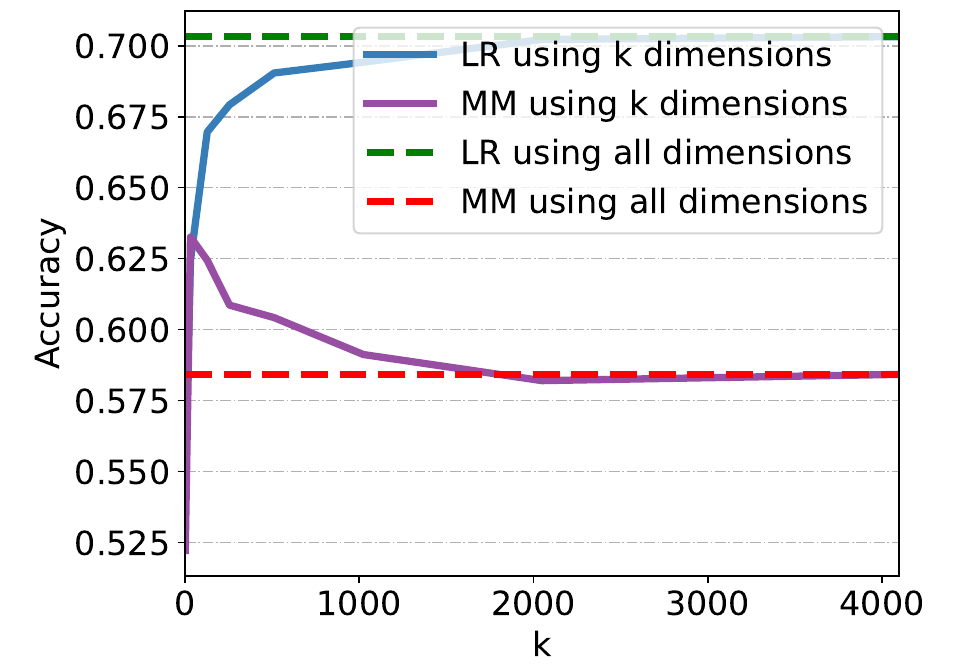}}
	\subfloat[TruthfulQA]{
		\includegraphics[scale=0.22]{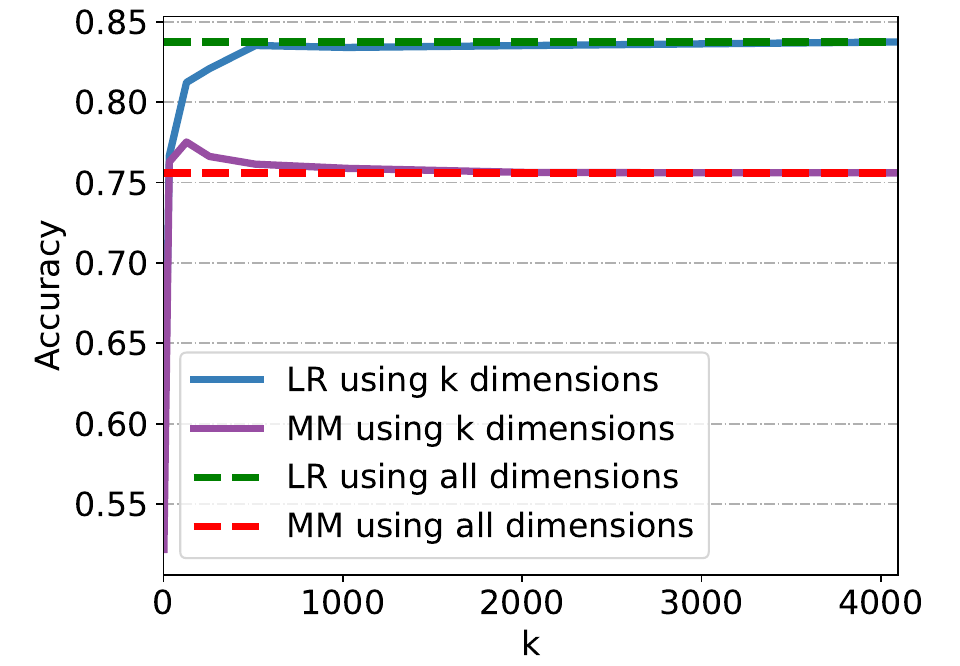} }
	\subfloat[Sciq]{
        \hspace{-0.3cm}
		\includegraphics[scale=0.22]{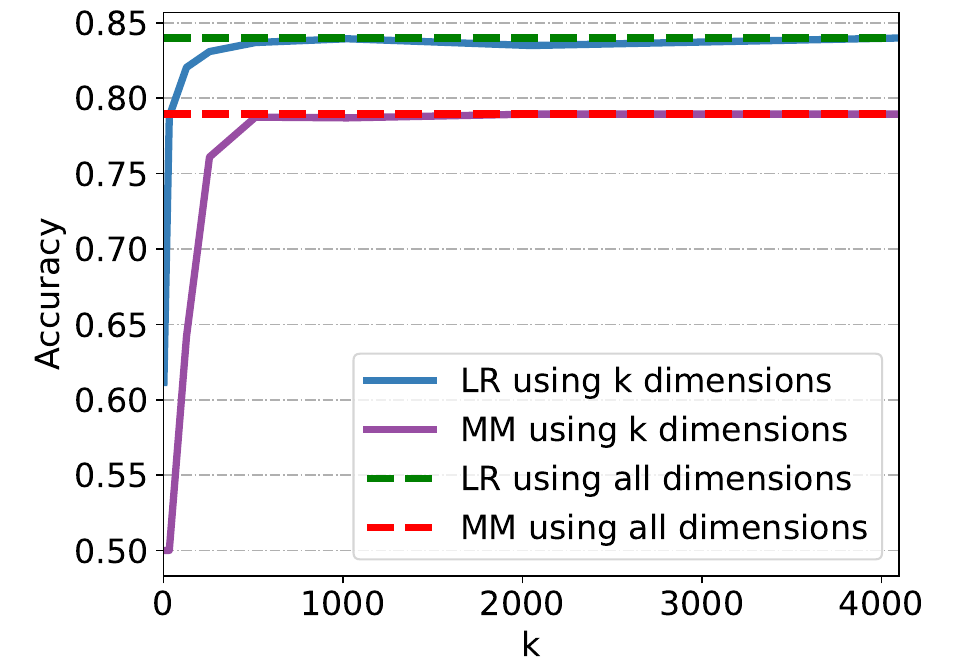} }
	\subfloat[Creak]{
        \hspace{-0.3cm}
		\includegraphics[scale=0.22]{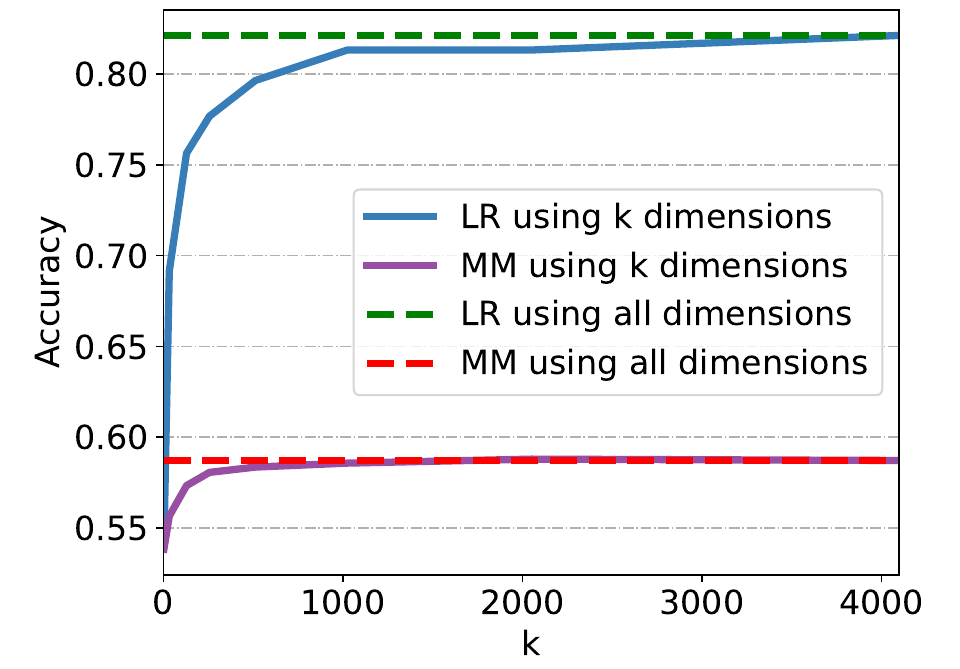} }
	\caption{Some other sparsity observations of layer residual activations on different tasks using the logistic regression (LR) and the mass mean (MM) probe. 
 }
	\label{fig-sparsity-show-layer} 
\end{figure*}

Besides, we also explore the sparsity on layer residual activations. Following the same experiment setting, the result is shown in Figure~\ref{fig-sparsity-show-layer}. We observe that using less than 1024 neurons can achieve comparable performance than using all 4096 neurons.

\section{Details on Hyperparameters Tuning}
\label{sec:appendix-hyperpara-tune}
We have two tunable hyperparameters for the Probe method: $num$ for the number of selected representations and $k$ for the compressed dimensions for every representation.
We note that we select $num$ positions according to  each validation split. However, we tune the $k$ and $num$ hyperparameters on the test splits of training tasks, that we select the hyperparameters that achieves highest accuracy on the test splits of training tasks. Therefore, it's important to note that we never tune the hyperparameters on validation or test splits of the test tasks.

The range of $k$ is always 64, 128.
When conducting experiment training the probe on single dataset in \textsection\ref{ood-failure}, the range of $num$ is 1, 2, 4, 10, 20, 30, 40, 60, 120. 
When conducting experiment training on all training tasks in \textsection\ref{main-exp}, \textsection\ref{exp-other-model}, and the study of training splits size in \textsection\ref{analysis}, the range of $num$ is 1,2,4.
When training the probe on the varying number of training tasks in \textsection\ref{head-vs-layer}: the experiment of comparing attention head and layer residual activations and the experiment of varying the number of training datasets, the $num$ is still selected from 1, 2, 4, 10, 20, 30, 40, 60, 120. However, we control the upperbound for $num$ as $160/t$, where $t$ is the number of datasets used training, to make sure a consistent upper bound for the overall selected positions when varying the training tasks.

\section{Experiment Details Setting}
\label{sec:appendix-experiment}




\subsection{Probes Fail to Generalize}
\label{detail-fail}
To evaluate in-distribution performance on the TruthfulQA dataset, we implement a 5-shot Probability baseline. This involves selecting five data samples from the TruthfulQA dataset to serve as demonstrations. We then measure the normalized probability and determine a threshold that maximizes accuracy on the TruthfulQA training split. Similarly, we apply the 5-shot approach when implementing the Self-Eval baseline.
For out-of-distribution (OOD) testing, we employ the Self-Eval baseline in a 0-shot setting, which does not rely on any prior examples. The detailed results for the OOD test are presented in Table~\ref{tab:tqa-failure}.

\begin{table*}[h] 
\centering
\resizebox{1.5\columnwidth}{!}{
\begin{tabular}{l ccc cc ccc c}
\toprule
\multirow{2}{*}{\textbf{Method}} & \multicolumn{3}{c}{\textbf{Short Answer Close Book QA}}&
\multicolumn{2}{c}{\textbf{Summarization}} &
\multicolumn{3}{c}{\textbf{Sentence Completion}} & \multirow{2}{*}{\textbf{Average}}\\
& NQ & Trivia QA & SciQ & XSum & CNN DM & Story Cloze & Hellaswag & CoPA &  \\
\midrule
Probe (LR) & 60.40 & 54.70 & 51.25 & 58.06 & 52.30 & 62.26 & 50.02 & 46.50 & 54.44 \\
Probe (MM) & 51.70 & 50.42 & 49.80 & 53.06 & 49.56 & 50.19 & 50.98 & 50.00 & 50.71 \\
Self-Eval 0-shot & 58.40 & 68.74 & 82.25 & 67.00 & 65.98 & 53.69 & 51.90 & 58.50 & \textbf{63.31} \\
FT & 62.38 & 68.44 & 62.90 & 52.56 & 51.26 & 53.55 & 50.98 & 50.00 & 56.51\\
Random & 50.00 & 50.00 & 50.00 & 50.00 & 50.00 & 50.00 & 50.00 & 50.00 & 50.00\\
\bottomrule
\end{tabular}
}
\caption{Probe trained on TruthfulQA, Self-Eval 0-shot baseline and FT (finetuning) method hallucination detection accuracy (\%) on OOD test sets.}
\label{tab:tqa-failure}
\end{table*}

\subsection{Main Experiments}
\label{detail-main-results}
Basically, we follow the principle that select few shot demonstrations or threshold from the same dataset (in-domain), a different dataset within the same  task (cross-domain), and a dataset from a different task (cross-task).

\paragraph{Probability baseline of cross-task}

When testing on the Short Answer Close Book QA task, considering Hotpot QA's~\citep{hotpotqa} format or type is close to the task, we rely on the Hotpot QA for the few shot demonstrations and threshold. To be specific, we first conduct the 5-shot Probability experiments on the Hotpot QA and then scan to find the threshold that achieves the highest accuracy on Hotpot QA's training split. Using the threshold and 5 correct demonstration from Hotpot QA, we then evaluate on the Short Answer Close Book QA task.
When testing on the Summarization task, we use 3 correct demonstrations from WEBNLG~\citep{gardent2017creating} dataset and also use the threshold that makes WEBNLG training split highest accuracy.
When testing on the Sentence Completion task, considering the tasks all are multi-choice QA, we use 5 correct ARC easy~\citep{allenai:arc} as demonstrations and use the ARC easy's threshold.

\paragraph{Probability baseline of cross-domain}

In the Short Answer Close Book QA task, we use Trivia QA~\citep{2017arXivtriviaqa} for demonstrations and threshold when testing SciQ~\citep{welbl2017crowdsourcing} and NQ~\citep{kwiatkowski2019natural} and we use SciQ for or demonstrations and threshold when testing Trivia QA.
In the the Summarization task, considering the summarization tasks's data is too long that not appropriate selected as few shot demonstrations, we still use WEBNLG as demonstrations. When testing XSum~\citep{Narayan2018DontGM}, we use the threshold that makes CNN Daily Mail's~\citep{hermann2015teaching, see2017get} training set highest accuracy during 3 shot Probability (demonstrations from WEBNLG) experiment. When testing CNN Daily Mail, we use threshold from XSum.
In the Sentence Completion task, when testing story cloze~\citep{mostafazadeh2017lsdsem} and HellaSwag~\citep{zellers2019hellaswag}, we use 5 shot demonstrations and threshold from CoPA~\citep{roemmele2011choice}. When testing CoPA, we use 5 shot demonstrations and threshold from story cloze.

\paragraph{Probability baseline of in-domain}
We all use threshold that makes its training split highest accuracy. We use few shot demonstrations from its training set except Summarization task that we still use WEBNLG~\citep{gardent2017creating} since the data is too long.

\paragraph{Self-Eval baseline of cross-task} 

When testing the Short Answer Close Book QA task, we use 5 data (labeled with Correct or Wrong) from Hotpot QA~\citep{hotpotqa} as few shot demonstrations.
When testing the Summarization task, as mentioned above that the data is so long that the model is hard to follow our aim to judge "Correct" or "Wrong", we here use 0 shot prompt like "Is the answer correct or wrong?\textbackslash nIt is"
When testing the Sentence Completion task, we use 5 data (labeled with Correct or Wrong) from ARC easy.

\paragraph{Self-Eval baseline of cross-domain} 

In the Short Answer Close Book QA task, we use Trivia QA~\citep{2017arXivtriviaqa} for demonstrations when testing SciQ and NQ and we use SciQ for demonstrations and threshold when testing Trivia QA.
In the  Summarization task, we still use 0 shot prompt.
In the Sentence Completion task, when testing story cloze~\citep{mostafazadeh2017lsdsem} and HellaSwag~\citep{zellers2019hellaswag}, we use 5 shot demonstrations from CoPA~\citep{roemmele2011choice}. When testing CoPA, we use 5 shot demonstrations  from story cloze.

\paragraph{Self-Eval baseline of in-domain}
We use demonstrations selected from its training set except Summarization that we still use 0 shot.

\paragraph{Finetune model setting}
We construct data samples using the prompt like

``Please determine whether the following answer is correct.

[data]

It is correct/wrong.
''

We use these constructed data to full finetuning the model and use same prompt and constrain model generate from "correct" and "wrong" two tokens when evaluating.
When training datasets contain fewer than 14 tasks, we use a learning rate of 2e-5 and train the model for 3 epochs. In contrast, when training datasets contain more than 14 tasks, we use a learning rate of 2e-5 and train the model for only 1 epoch. 

\section{Experiment Details for Training on Attention Head and Layer Activations}
\label{sec:appendix-preliminary}

In our study, we have explored training probes using the layer residual activations and attention head outputs, finding that probes trained on layer activations consistently underperform attention head outputs. 

We conduct the cross-task experiments with varying number of training datasets, 4 datasets, 8 datasets, 12 datasets respectively. When training the probes on attention head outputs, following the hyperparameters range: $k$ can be 64 or 128, $num$ can be selected from 1, 2, 4, 10, 20, 30, 40, 60, 120, but maintain the consistent upper bound $160/t$, where $t$ is the number of training datasets. For training probes on layer residual activations, we also utilize the same framework, including $k$ and $num$ two hyperparameters, where $k$ can be 1024, 4096 and $num$ fixed at 1, reflecting the limited selection options available for layers.

\end{document}